%% file: main.tex
\def\isarxiv{1} 

\ifdefined\isarxiv
\documentclass[11pt]{article}

\usepackage[numbers]{natbib}

\else
\documentclass{article}

\usepackage{icml2025}

\usepackage[textsize=tiny]{todonotes}

\icmltitlerunning{Video Latent Flow Matching: Optimal Polynomial Projections for Video Interpolation and Extrapolation}

\fi

\usepackage{amsmath}
\usepackage{amsthm}
\usepackage{amssymb}
\usepackage{algorithm}
\usepackage{algpseudocode}
\usepackage{graphicx}
\usepackage{grffile}
\usepackage{wrapfig,epsfig}
\usepackage{url}
\usepackage{xcolor}
\usepackage{epstopdf}

\usepackage{microtype}
\usepackage{graphicx}
\usepackage{subfig}
\usepackage{booktabs} 

\usepackage{hyperref}

\usepackage{bbm}
\usepackage{dsfont}

\allowdisplaybreaks

\ifdefined\isarxiv

\usepackage{tikz}
\usepackage{hyperref}  
\hypersetup{colorlinks=true,citecolor=blue,linkcolor=blue} 
\usetikzlibrary{arrows}
\usepackage[margin=1in]{geometry}

\fi
 
\graphicspath{{./figs/}}

\theoremstyle{plain}
\newtheorem{theorem}{Theorem}[section]
\newtheorem{lemma}[theorem]{Lemma}
\newtheorem{definition}[theorem]{Definition}

\newtheorem{assumption}[theorem]{Assumption}

\newtheorem{fact}[theorem]{Fact}

\newtheorem{example}[theorem]{Example}

\newcommand{\wh}{\widehat}
\newcommand{\wt}{\widetilde}

\newcommand{\R}{\mathbb{R}}

\renewcommand{\d}{\mathrm{d}}

\renewcommand{\hat}{\wh}

\DeclareMathOperator*{\E}{{\mathbb{E}}}

\DeclareMathOperator{\diag}{diag}

\makeatletter
\newcommand*{\RN}[1]{\expandafter\@slowromancap\romannumeral #1@}
\makeatother

\usepackage{lineno}

\begin{document}

\ifdefined\isarxiv

\date{}

\title{Video Latent Flow Matching: Optimal Polynomial Projections for Video Interpolation and Extrapolation}

\author{
Yang Cao\thanks{\texttt{ycao4@wyomingseminary.org}. Wyoming Seminary.}
\and
Zhao Song\thanks{\texttt{magic.linuxkde@gmail.com}. Simons Institute for the Theory of Computing, University of California, Berkeley.}
\and
Chiwun Yang\thanks{\texttt{christiannyang37@gmail.com}. Sun Yat-sen University.}
}

\else

\twocolumn[
\icmltitle{Video Latent Flow Matching:\\Optimal Polynomial Projections for Video Interpolation and Extrapolation}

\begin{icmlauthorlist}
\icmlauthor{Yang Cao}{sem}
\icmlauthor{Zhao Song}{simons}
\icmlauthor{Chiwun Yang}{sun}
\end{icmlauthorlist}

\icmlaffiliation{sem}{Wyoming Seminary. Kingston, PA 18704, USA}
\icmlaffiliation{simons}{Simons Institute for the Theory of Computing, University of California, Berkeley. Berkeley, CA 94720, USA}
\icmlaffiliation{sun}{Sun Yat-sen University. Guangzhou, Guangdong, China}

\icmlcorrespondingauthor{Yang Cao}{ycao4@wyomingseminary.org}
\icmlcorrespondingauthor{Zhao Song}{magic.linuxkde@gmail.com}
\icmlcorrespondingauthor{Chiwun Yang}{christiannyang37@gmail.com}

\icmlkeywords{Deep Learning, Video Generation, Diffusion Models, Flow Matching}

\vskip 0.3in
]
\printAffiliationsAndNotice{}

\fi

\ifdefined\isarxiv
\begin{titlepage}
  \maketitle
\begin{abstract}
\input{0_abstract}

\end{abstract}
\thispagestyle{empty}
\end{titlepage}

{\hypersetup{linkcolor=black}
\tableofcontents
}
\newpage

\else

\begin{abstract}
\input{0_abstract}
\end{abstract}

\fi

\input{1_intro}

\input{2_related_work}

\input{3_preli}

\input{4_vlfm}

\input{5_theory}

\input{6_exps}

\input{7_conclusion}

\ifdefined \isarxiv
\else
\input{9999_impact}
\section*{Acknowledgments}
\bibliography{ref}
\bibliographystyle{icml2025}
\fi

\newpage
\onecolumn
\appendix

\input{8_app_more_result}

\input{9_app_preli}

\input{10_app_vlfm}

\input{11_app_dit}

\input{12_app_interpolation}

\ifdefined \isarxiv
\bibliography{ref}
\bibliographystyle{alpha}
\fi



\end{document}

%% file: 0_abstract.tex
This paper considers an efficient video modeling process called Video Latent Flow Matching (VLFM). Unlike prior works, which randomly sampled latent patches for video generation, our method relies on current strong pre-trained image generation models, modeling a certain caption-guided flow of latent patches that can be decoded to time-dependent video frames. 
We first speculate multiple images of a video are differentiable with respect to time in some latent space. Based on this conjecture, we introduce the HiPPO framework to approximate the optimal projection for polynomials to generate the probability path. Our approach gains the theoretical benefits of the bounded universal approximation error and timescale robustness. Moreover, VLFM processes the interpolation and extrapolation abilities for video generation with arbitrary frame rates. We conduct experiments on several text-to-video datasets to showcase the effectiveness of our method.

%% file: 1_intro.tex
\ifdefined\isarxiv
\else
\vspace{-8mm}
\fi

\begin{figure*}[!ht]
\begin{center}
\centering
    \includegraphics[width=0.9\textwidth]{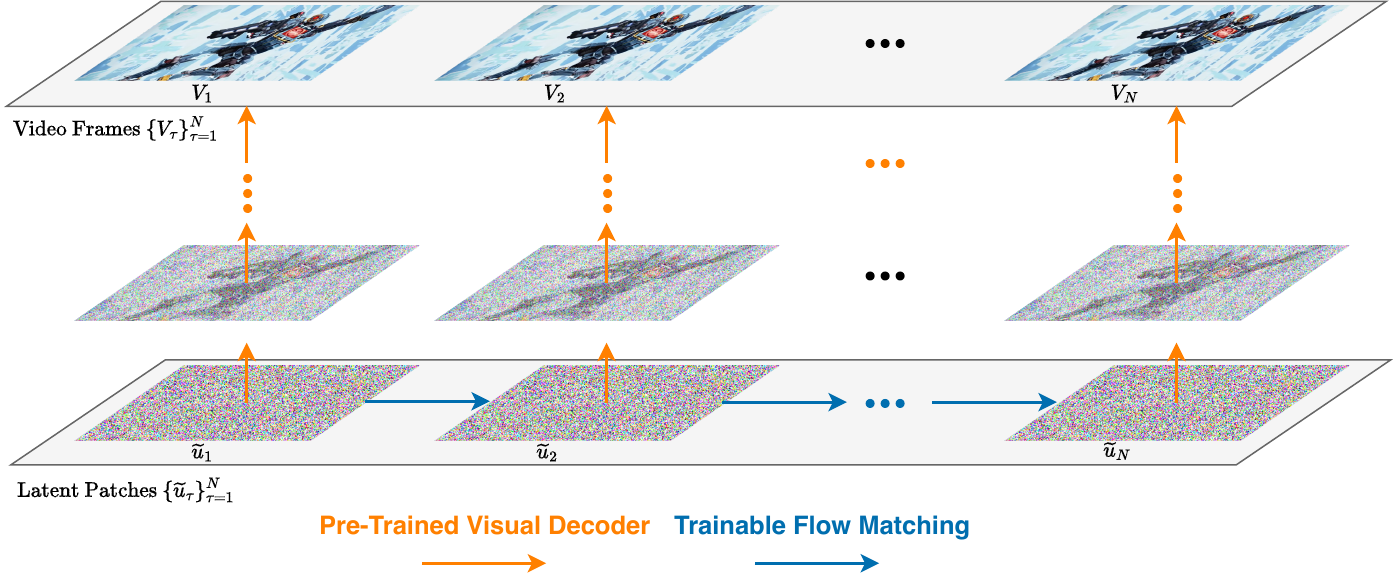}
\end{center}
\caption{Illustration of the working mechanism behind {\it Video Latent Flow Matching}.}
\label{fig:vlfm}
\end{figure*}

\section{Introduction}

The rise of generative models has already demonstrated excellent performance in various fields like image generation \cite{scs+22,rbl+22}, text generation \cite{aaa23,djp+24,lfx+24}, video generation \cite{bph+24,zpy+24,jsl+24,tjy+24}, etc. \cite{s23}. Among them, some of the most popular algorithms - Flow Matching \cite{lcb+22,lgl22}, Diffusion \cite{hja20,sme20} and VAEs \cite{kw13}, perform surprise generative capabilities, however, requiring comprehensive computational resources for training. In particular, this efficiency drawback harms the development of more successful text-to-video modeling \cite{bph+24}, becoming a frontier challenge in the field of generative modeling.

The prior works about the generation from textual descriptions to realistic and coherent videos usually involve two strong pre-trained networks \cite{hsg+22,zpy+24}. One encodes input captions to rich embedding representations, and another one decodes from sequences of latent patches (also considered as Gaussian noise) under the guidance of text embedding representations. Although variants based on such modeling processes are already showing some fine initial results, the necessity of training on large-scale models and datasets leads these studies to be undemocratic \cite{bph+24,ktz+24}. In response to this issue, the motivation of this paper is to design a novel algorithm to simplify the process of text-to-video modeling.

In this paper, we propose \emph{Video Latent Flow Matching} (VLFM), which relies on the most advanced pre-trained image generation models (we call visual decoder in the range of this paper) for their extension in the field of text-to-video generation. In detail, we first introduce a deterministic inversion algorithm \cite{sme20,lcb+22,lgl22} to the visual decoder and apply this inversion operation to the frames of all videos, obtaining a sequence including initial latent patches from each video. Thus, the base of this paper is that a sequence of latent patches is a time-dependent and caption-conditional flow, so-called {\it Video Latent Flow}. Therefore, we use Flow Matching \cite{lcb+22,lgl22} to model it. 

Especially, we emphasize four advantages of our VLFM:
\begin{itemize}
    \item {\bf Modeling efficiency. } The modeling of VLFM only needs to fit ${\sf N}$ flows where ${\sf N}$ is the size of the training dataset. This computational requirement is close to training a text-to-image model.
    \item {\bf Optimal polynomial projections. } We use discrete HiPPO LegS to generate the time-dependent flow with provable optimal polynomial projections. The approximating error decreases with the enlarging order of polynomials.
    \item {\bf Arbitrary frame rate. } The reason for applying Flow Matching instead of other approaches is that it allows solving ODE with arbitrary time $t$. This further leads to precise video generation with high frame rates.
    \item {\bf Interpolation and extrapolation. } Besides, VLFM is suitable for interpolation and extrapolation for high-precision video recovery and generation since its generalization performance is confirmed in our theoretical part.
\end{itemize}
\ifdefined\isarxiv
\else
\vspace{-3mm}
\fi

In summary, we make the following contributions:
\begin{itemize}
    \item We give this paper's preliminary as a theoretical background with several mild assumptions in Section~\ref{sec:preli}. Hence, we state the derivation of our VLFM in Section~\ref{sec:vlfm}, which introduces the HiPPO framework to online approximate the sequence of latent patches.
    \item The theoretical benefits of VLFM are shown in Section~\ref{sec:theory}. We first utilize the universal approximation theorem of Diffusion Transformer (DiT) to ensure an appropriate learner for modeling. The approximation bound then is guaranteed. We also discuss how our VLFM processes interpolation and extrapolation to real-world videos with an upper bound on error and its timescale robustness.
    \item We validate our approach by conducting extensive experiments in Section~\ref{sec:exp}. Our model leverages DiT-XL-2 and is trained on a diverse collection of seven large-scale video datasets, including OpenVid-1M, MiraData, and videos from Pixabay. The results demonstrate strong performance in text-to-video generation, interpolation, and extrapolation, achieving robust and reliable outputs with significant potential for real-world video applications.
\end{itemize}

%% file: 2_related_work.tex
\section{Related Work}

This section briefly reviews three topics that are closely related to this work: Text-to-Video Generation, Flow Matching, and Theory in Transformer-Based Models. 

{\bf Text-to-Video Generation.}
Text-to-video generation \cite{sph+22, vjp22, brl+23} is a specialized form of conditional video generation that aims to synthesize high-quality videos from textual descriptions. Recent advancements in this field have predominantly leveraged diffusion models \cite{ssk+20,hja20}, which iteratively refine video frames by learning to denoise samples from a normal distribution. This approach has proven effective in generating coherent and visually appealing videos.
Training strategies for text-to-video models vary widely. One common approach involves adapting pre-trained text-to-image models by incorporating temporal modules, such as temporal convolutions and attention mechanisms, to establish inter-frame relationships \cite{gnl+23, azy+23, sph+22, gwz+23, gyr+23}. For instance, PYoCo \cite{gnl+23} introduced a noise prior technique and utilized the pre-trained eDiff-I model \cite{bnh+22} as a starting point. Alternatively, some methods build on Stable Diffusion \cite{rbl+22}, leveraging its accessibility and pre-trained capabilities to expedite convergence \cite{brl+23,zwy+22}. However, this approach can sometimes result in suboptimal outcomes due to the inherent distributional differences between images and videos. Another strategy involves training models from scratch on combined image and video datasets \cite{hcs+22}, which can yield superior results while requiring intensive computationally.
\ifdefined\isarxiv
\else
\vspace{-2mm}
\fi

{\bf Flow Matching.}
Flow Matching has emerged as a highly effective framework for generative modeling, demonstrating significant advancements across various domains, including video generation. Its simplicity and power have been validated in large-scale generation tasks such as image \cite{ekb+24}, video \cite{pzb+24,jsl+24}, speech \cite{lvs+24},  audio \cite{vsl+23}, proteins \cite{hvf+24}, and robotics \cite{bbd+24}. Flow Matching originated from efforts to address the computational challenges associated with Continuous Normalizing Flows (CNFs), where early methods struggled with simulation inefficiencies \cite{rgnl21,bcba+22}. Modern Flow Matching algorithms \cite{lcb+22,lgl22,av22,nbsm23,hbc23,tfm+23} have since evolved to learn CNFs without explicit simulation, significantly improving scalability. Recent innovations, such as Discrete Flow Matching \cite{cyb+24,grs+24}, have further expanded the applicability of this framework, making it a versatile tool for generative tasks.

{\bf Theory in Transformer-Based Models.}
Transformers have become a cornerstone in AI and are widely used in different areas, especially in NLP (Natural Language Process) and CV (Computer Vision). However, understanding the Transformers from a theoretical perspective remains an ongoing challenge. Several works have explored the theoretical foundations and computational complexities of the Transformers 
\ifdefined\isarxiv
\cite{tby+19,zhdk23,bsz23,as24,syz24,cll+24_rope,hlsl24,mosw22,szz24,als19_icml,dhs+22,bpsw21,syz21,als+23,dsxy23,kls+24,lswy23,gls+24}, 
\else
\cite{tby+19,zhdk23,bsz23,as24,syz24,cll+24_rope,hlsl24,mosw22,szz24,als19_icml,dhs+22,bpsw21,syz21,als+23,dsxy23}
\fi
focusing on areas such as efficient Transformers \cite{hjk+23,smn+24,szz+21,lll21,lls+24_conv,lssz24_tat,lss+24,llss24_sparse,lls+24_prune,cls+24,lls+24_io,hwsl24,hwl24,hcl+24,whhl24,hyw+23,as24_arxiv,gswy23}, optimization \cite{dls23, csy24}, and the analysis of emergent abilities \cite{bmr+20,wtb+22,al23,j23,xsw+23,lls+24_grok,xsl24,cll+24,lss+24_relu,hwg+24,wsh+24, dsy24_b}. Notably, \cite{zhdk23, dsy24_a} introduced an algorithm with provable guarantees for approximation of Transformers, \cite{kwh23} proved a lower bound for Transformers based on the Strong Exponential Time Hypothesis, and \cite{as24} provided both an algorithm and hardness results for static Transformers computation.
\ifdefined\isarxiv
\else
\vspace{-3mm}
\fi

%% file: 3_preli.tex
\section{Preliminary}\label{sec:preli}

In this section, we formalize the background of this paper. We first introduce how we invert video frames into some latent space using the strong pre-trained visual decoder in Section~\ref{sub:data}. We state the definition of data and assumption in Section~\ref{sub:assumptions}. Section~\ref{sub:problem_def} defines the main problem we aim to address in this paper. We use integer $s$ to denote the order of polynomials. The dimensional number of the text embedding vector is given by integer $\ell$.

\subsection{Inverting Video Frames to Latent Patches}\label{sub:data}

\paragraph{Notations.} We use $D$ to denote the flattened dimension of real-world images. We use $d$ to represent the dimension of latent patches. We introduce $d_0$ as the dimension of Diffusion Transformers. We utilize $V: [0, T] \rightarrow \R^D$ to denote a video with $T$ duration, where $T$ is the longest time for each video. We omit $\nabla_t a(t)$ and $a'(t)$ to denote taking differentiation to some function $a(t)$ w.r.t. time $t$.

\paragraph{Visual decoder.} Here we denote the visual decoder ${\cal D}: \R^d \rightarrow \R^D$ satisfies that: For any flattened image $V \in \R^D$, there is a unique $u \in \R^d$ such that ${\cal D}(u) = V$. Then we say ${\cal D}$ is bijective. Denote the reverse function of ${\cal D}$ as ${\cal D}^{-1}: \R^D \rightarrow \R^d$. Note that this visual decoder ${\cal D}$ could be considered as any generative algorithm practically, e.g. LDM \cite{rbl+22}, DDIM \cite{sme20} and VAE \cite{k13}. We thus implement an inversion algorithm to invert video frames to latent patches \cite{mha+23}. In particular, we define these latent patches here, which depend on the detailed visual decoder. We consider these latent patches following arbitrary distribution.

We abuse the notation $u: [0, T] \rightarrow \R^d$ to denote a sequence of latent patches of a video $V$. In detail, we define: $u_t := {\cal D}^{-1}(V_t)$ for any $t \in [0, T]$.

\paragraph{Discretization for cases of real-world data.} We denote $\Delta t$ as the minimal time unit of measurement in the real world (Planck time). Hence, a video $V$ with $T$ duration can be divided into at most $\frac{T}{\Delta t}$ frames. We use matrix $\wt{V} \in \R^{\frac{T}{\Delta t} \times D}$ to denote the compact form of discretized video. We use $\Phi \in \{0, 1\}^{\frac{T}{\Delta t} \times N}$ for $N \leq \frac{T}{\Delta t}$ to denote the corresponding observation matrix due to the real-world consideration, especially $\Phi^\top {\bf 1}_{\frac{T}{\Delta t}} = {\bf 1}_{N}$. Then the practical form of latent patches is given by:
\begin{align}\label{eq:u_tau:informal}
    \wt{u}_\tau := {\cal D}^{-1}([ \Phi \wt{V} ]_\tau) \in \R^{d}, \forall \tau \in [N].
\end{align}

\subsection{Data and Assumptions}\label{sub:assumptions}

\paragraph{Caption-video data pairs.} Given a video distribution ${\cal V}$, we introduce a text embedding state distribution ${\cal C}$ that maps one-to-one to ${\cal V}$. Then for any video data $V \sim {\cal V}$, $c \in \R^{\ell}$ is denoted as the corresponding caption embedding state vector. We use ${\cal V}_c$ to denote the distribution that contains video and embedding vector, such that $(V, c) \sim {\cal V}_c$.

\paragraph{Assumptions.} Here we list several mild assumptions in this paper, such that:
\begin{itemize}
    \item {\bf $k$-differentiable latent patches $u$. } We assume $u: [0, T] \rightarrow \R^d$ is a differentiable function with order $k$. 
    \item {\bf Lipschitz smooth visual decoder function ${\cal D}$. } We assume the visual decoder function ${\cal D}$ is $L_0$-smooth for constant $L_0 > 0$. Formally, it is: $\| {\cal D}(x) - {\cal D}(y) \|_2 \leq L_0 \cdot \| x - y\|_2, \forall x, y \in \R^d$.
    \item {\bf Bounded entries of $u$.} For each entry in latent patches $u$, we assume it is smaller than a upper bound $U$ for some constant $U > 0$.
    \item {\bf Caption-to-latency function. } For any video-caption data $(V, c) \sim {\cal V}_c$, there exists a function ${\cal M}: [0, T] \times \R^\ell \rightarrow \R^D$ satisfies $V_t = {\cal M}_t(c)$. 
\end{itemize}

\subsection{Problem definition: Modeling Text-to-Latency Data from Discretized video}\label{sub:problem_def}

In this paper, we consider the video modeling problems as follows:
\begin{itemize}
    \item 
    Given a video-caption pair $({\cal V}, c) \sim {\cal V}_c$, we obtain the data $\wt{u}_\tau \in \R^d, \forall \tau \in [N]$ via Eq.~\eqref{eq:u_tau:informal}, we aim to find a algorithm that inputs a time $t \in [0, T]$ and encoded text state vector $c \in \R^{\ell}$ and output a predicted latent patch $\hat{u}_t \in \R^d$, it satisfies: 
    \begin{align}\label{eq:main}
        \| {\cal D}(\hat{u}_t) - V_t \|_p \leq \epsilon.
    \end{align}
    Here we denote the error $\epsilon \ge 0$ and some $\ell_p$ norm metric.
\end{itemize}

\paragraph{Connecting the main problem to interpolation and extrapolation.} Since the frames number $N$ of obtained video data may be greatly smaller than $T/\Delta t$. Recovering the continuous video data $T$ as completely as possible (both interpolation and extrapolation) would also be our goal in the range of this paper. Theoretically, we see such interpolation and extrapolation as one: given a discrete video data $\Phi \wt{V} \in \R^{N \times D}$, the sequence of latent patches is $\wt{u} = [\wt{u}_1^\top, \wt{u}_2^\top, \cdots, \wt{u}_N^\top] \in \R^{N \times d}$ using Eq.~\eqref{eq:u_tau:informal}. The text embedding state vector $c \in \R^{\ell}$ could be ensured by some video-to-caption methods. Our target is to find an algorithm that inputs $\wt{u}$ and outputs $\hat{u}_{\tau}, \tau \in [T/\Delta t]$ that meets the requirement: $\| {\cal D}(\hat{u}_\tau) - \wt{V}_\tau \|_p \leq \epsilon$ for error $\epsilon \ge 0$ and some $\ell_p$ norm metric.

%% file: 4_vlfm.tex
\section{Video Latent Flow Matching} \label{sec:vlfm}

In this section, we propose Video Latent Flow Matching (VLFM) in response to the main problem in Section~\ref{sub:problem_def}. Especially, we briefly review the HiPPO (high-order polynomial projection operators) framework \cite{gde+20} in Section~\ref{sub:hippo}. We state the derivation of our VLFM based on the popular flow matching approach \cite{lcb+22} in Section~\ref{sub:vlfm}. Finally, we define the training objective of the VLFM for efficient video modeling in Section~\ref{sub:training_objective}.

\subsection{HiPPO Framework and LegS State Space Model}\label{sub:hippo}

Given an input function $f(t) \in \R$ for $t \ge 0$, we use $f_{\leq t}$ to denote the cumulative history of $f(t)$ for every time $t \ge 0$. We choose integer $s \ge 1$ as the order of approximation. Then, any $s$-dimensional subspace ${\cal G}$ of this function space is a suitable candidate for the approximation. Given a time-varying measure family $p(t)$ supported on $(-\infty, t)$, a sequence of basis functions ${\cal G} = {\rm span}\{ g_{i}(t) \}_{i=1}^s$. HiPPO \cite{gde+20} defines an operator that maps $f$ to the optimal projection coefficients $c: \R_{\ge 0} \rightarrow \R^s$, such that:
\begin{align*}
    g(t) := & ~ \arg \min_{g \in {\cal G}} \| f_{\leq t} -  g \|_{p(t)}, \\
    g(t) = & ~ \sum_{i=1}^s c_i(t) \cdot g_i(t).
\end{align*}
We focus on the case where the coefficients $c(t)$ have the form of a linear ODE satisfying $\nabla c(t) = A(t) c(t) + B(t) f(t)$ for some $A(t) \in \R^{s \times s}$ and $B(t) \in \R^{s \times 1}$. This equation is now also known as the state space model (SSM) in many works \cite{kds+15,aia+22,gd23,dg24,zlz+24,xyy+24,mlw24,rx24,sld+24}.

{\bf Discrete HiPPO-LegS.} The setting of HiPPO-LegS defines the update rule of SSM and the discrete version of $A$ and $B$ matrices, which are $c_{\tau + 1} = (I_s - \frac{A}{\tau}) c_\tau + \frac{1}{\tau} B f_\tau$ and:
\begin{align*}
    A_{i_1, i_2} & ~ = \begin{cases}
        \sqrt{(2i_1 + 1)(2i_2 + 1)}, & \text{if $i_1 > i_2$} \\
        i_1 + 1, & \text{if $i_1 = i_2$} \\
        0, & \text{if $i_1 < i_2$}
    \end{cases}, \\
    B_{i_1} & ~ = \sqrt{2i_1 + 1}, \forall i_1, i_2 \in [s].
\end{align*}

\subsection{Conditional Video Latent Flow}\label{sub:vlfm}

Here we emphasize the core idea of VLFM is to approximate a continuous video distribution from limited discrete video frames data utilizing the optimal high-order polynomial approximation. 

Given a video-caption distribution ${\cal V}_c$, then for any video-caption data pair $(V, c) \sim {\cal V}_c$, we obtain the data $\wt{u}_\tau \in \R^d, \forall \tau \in [N]$ via Eq.~\eqref{eq:u_tau:informal}.
We aim to define a time-dependent flow $\psi_t(\wt{u})$ that takes inputs $\wt{u}$ and time $t$, and could match $\widehat{u}_\tau$ for all time $\tau \in [N]$. Since $\widehat{u}$ is discrete, HiPPO-LegS will be the best solution to approximate the continuous data. We define the \emph{Video Latent Flow} as:
\begin{align}\label{eq:psi}
    \psi_t(\wt{u}) := \sigma_t( \wt{u} ) \cdot z + \mu_t(\wt{u}) \in \R^d,
\end{align}
where $t \in [0,T]$ and $z \sim \mathcal{N}(0, I_d)$, $\sigma: [0, T] \times \R^{N \times d} \rightarrow \R_{> 0}$ denotes the time-dependent standard deviation, where $\sigma_0 ( \wt{u} ) = 1$, and $\sigma_{\frac{T}{N} \cdot \tau}( \wt{u} ) = \sigma_{\min}$, for all $\tau \in [N]$ ; $\mu: [0, T] \times \R^{N \times d} \rightarrow \R^d$ denotes the time-dependent mean of Gaussian distribution, where $\mu_0(\wt{u}) = {\bf 0}_d$, $\mu_{\frac{T}{N} \cdot \tau}(\wt{u}) = \wt{u}_\tau,$ for all $\tau \in [N]$.

Especially, we define:
\begin{align*}
    \mu_t(\wt{u} ) := & ~  H_N g(t), \\
    H_{\tau + 1} := & ~ H_{\tau} (I_s - \frac{1}{\tau} A)^\top + \frac{1}{\tau} \wt{u}_\tau B^\top,
\end{align*}
where $g(t) := [\sqrt{\frac{1}{2}} P_0(t), \sqrt{\frac{3}{2}} P_1(t), \cdots, \sqrt{\frac{2s-1}{2}} P_{s-1}(t)]^\top $ $\in \R^{s}$, $P_i(t), \forall i \in [s]$ is Legendre polynomials. We initialize $H_0 := {\bf 0}_{d \times s}$.

Besides, having a large scalar $\alpha > 0$, we give:
\begin{align*}
    \sigma_t(\wt{u}) := (1 - \sigma_{\min}) \cdot [\sin^2( \pi \frac{N}{T} t ) +  \exp(-\alpha t) ] + \sigma_{\min}.
\end{align*}

\subsection{Training Objective}\label{sub:training_objective}

Here we define a model function $F_\theta: \R^d \times \R^\ell \times [0, T] \rightarrow \R^d$ with parameters $\theta$ to learn the conditional video latent flow $\psi_t(\wt{u})$ defined in Eq.~\eqref{eq:psi}. This function takes inputs of flow and time to predict the vector field. The training objective is based on the Flow Matching framework \cite{lcb+22}, which aims to minimize the distance between the model's prediction and the true derivative of the flow.

The training objective of VLFM is defined as the expectation of the square $\ell_2$ norm of the difference, which is:
\begin{align*}
    {\cal L}(\theta) := \E_{z, t, (V, c)}[\| F_\theta( \psi_t(\wt{u}), c, t ) - \frac{\d }{\d t} \psi_t(\wt{u}) \|_2^2],
\end{align*}
where $z \sim \mathcal{N}(0, I_d)$,  $t \sim {\sf Uniform}[0, T]$ and $(V, c) \sim {\cal V}_c$. By minimizing this objective, the model learns to approximate the vector field that transports the initial noise distribution to the distribution of video latent patches. Formally, we solve: $\min_{\theta} {\cal L}(\theta)$. 

{\bf Close-form solution.} Furthermore, the close-form solution could be easily obtained as follows:
\begin{theorem}
    The minimum solution for function $F_\theta$ that takes $z \sim N(0, I_d)$ and $t \sim {\sf Uniform}[0, T]$ is:
    \begin{align*}
        F_\theta(z, c, t) = \frac{\sigma_t'(\wt{u})}{\sigma_t(\wt{u})} (z - \mu_t(\wt{u})) + \mu_t'(\wt{u}).
    \end{align*}
\end{theorem}

\begin{proof}
    This proof follows from Theorem 3 in \cite{lcb+22}.
\end{proof}

%% file: 5_theory.tex
\section{Theory}\label{sec:theory}

This section provides several theoretical advantages of our VLFM. The approximation theory in this approach builds up based on using the Diffusion Transformer (DiT) \cite{px23}, which is a popular choice in previous empirical and theoretical part generative model works \cite{chzw23, hwsl24}, we briefly state its definitions in Section~\ref{sub:DiT}.

In addition, we provide the optimal polynomial projection guarantee and universal approximation theorem (with DiT) of VLFM in Section~\ref{sub:approx} to confirm its approximating ability. Besides, Section~\ref{sub:inter-extra_polation_theory} gives error bound of interpolation and extrapolation, and Section~\ref{sub:timescale_robustness} gives the supplementary property that VLFM's timescale robustness, which indicates its theoretical advantages.

\subsection{Diffusion Transformer (DiT)}\label{sub:DiT}

Diffusion Transformer \cite{px23} is a framework that utilizes Transformers \cite{vnn+17} as the backbone for Diffusion Models \cite{hja20,sme20}. Specifically, a Transformer block consists of a multi-head self-attention layer and a feed-forward layer, with both layers having a skip connection. 
We use ${\sf TF}^{h, m, r}: \R^{n \times d_0}\rightarrow \R^{n \times d_0}$ to denote a Transformer block.
Here $h$ and $m$ are the number of heads and head size in self-attention layer, and $r$ is the hidden dimension in feed-forward layer.
Let $X \in \R^{n \times d_0}$ be the model input. Then, we have the model output:
\ifdefined\isarxiv
\begin{align*}
    {\sf Attn}(X) := \sum_{i=1}^h {\sf Softmax}( X W_Q^i {W_K^i}^\top X^\top ) \cdot X W_V^i {W_O^i}^\top + X,
\end{align*}
\else
\begin{align*}
    & ~ {\sf Attn}(X) \\
    & ~ := \sum_{i=1}^h {\sf Softmax}( X W_Q^i {W_K^i}^\top X^\top ) \cdot X W_V^i {W_O^i}^\top + X,
\end{align*}
\fi
where the projection weights $W_K^i, W_Q^i, W_V^i, W_O^i \in \R^{d_0 \times m}$. Moreover,
\begin{align*}
    {\sf FF}(X) := \phi(X W_1 + {\bf 1}_n b_1^\top) \cdot W_2^\top + {\bf 1}_n b_2^\top + X.
\end{align*}
where  the projection weights $W_1, W_2 \in \R^{d_0 \times r}$, bias $b_1 \in \R^{r}, b_2 \in \R^{d_0}$, and $\phi$ is usually considered as the ReLU activated function.

In our work, we use Transformer networks with positional encoding $E\in\R^{n \times d_0}$. The transformer networks are then defined as the composition of Transformer blocks:
\begin{align*}
    {\cal T}_{P}^{h,m,r} = & ~ \{f_{{\cal T}}:\R^{ n \times d_0 }\rightarrow {\R^{n \times d_0}} \\
    & ~ \mid f_{{\cal T}}\text{ is a composition of blocks }{\sf TF}^{h,m,r}\text{'s}\}.
\end{align*}
For example, the following is a Transformer network consisting $K$ blocks and positional encoding
\begin{align*}
f_{{\cal T}}(X)= {\sf FF}^{(K)} \circ {\sf Attn}^{(K)} \circ  \cdots {\sf FF}^{(1)} \circ  {\sf Attn}^{(1)} (X+E).
\end{align*}

\subsection{Approximation via DiT}\label{sub:approx}

Before we state the approximation theorem, we define a reshaped layer that transforms concatenated input in flow matching into a length-fixed sequence of vectors. It is denoted as $R: \R^{d+\ell+1} \rightarrow \R^{n \times d_0}$. Therefore, in the following, we give the theorem utilizing DiT to minimize training objective ${\cal L}(\theta)$ to arbitrary error.

\begin{theorem}[Informal version of Theorem~\ref{thm:uat}]\label{thm:uat:informal}
    There exists a transformer network $f_{\cal T} \in {\cal T}_{P}^{2, 1, 4}$ defining function $F_\theta(z, c, t) := f_{\cal T}( R([z^\top, c^\top, t]^\top) )$ with parameters $\theta$ that satisfies ${\cal L}(\theta) \leq \epsilon$ for any error $\epsilon > 0$. 
\end{theorem}

\begin{proof}[Proof sketch of Theorem~\ref{thm:uat:informal}]
    Please refer to the proof of Theorem~\ref{thm:uat} for the detailed analysis.
\end{proof}

\subsection{Interpolation and Extrapolation}\label{sub:inter-extra_polation_theory}

Now, we theoretically discuss the approximating error of our VLFM in processing interpolation and extrapolation. It is considered a recovery of the original idea data from limited sub-sampled observations. This analysis is achieved by splitting the error into three parts, which are: 1) approximating error $\epsilon_1$ for HiPPO-LegS approximating the original data; 2) Gaussian error $\epsilon_2$ for the boundary of Gaussian vector $z$; 3) interpolation and extrapolation error $\epsilon_3$ that represents the training and predicting the difference between using original idea data $V$ and limited sub-sampled observations $\Phi \wt{V}$. We state the results as follows:
\begin{lemma}[Informal version of Lemma~\ref{lem:hippo_error}]\label{lem:hippo_error:informal}
    Denote failure probability $\delta \in (0, 0.1)$. Let the flow $\psi_t( \wt{u} )$ defined in Eq.~\eqref{eq:psi}. Denote $G := [g(\Delta t), g(2 \Delta t), \cdots, g(T)]^\top \in \R^{\frac{T}{\Delta t} \times s}$ and $\lambda^* := \lambda_{\min}(G) > 0$ as the minimum eigenvalue of $G$. Choosing $s = O(\frac{\Delta t}{T}\log((\frac{\Delta t}{T})^{1.5}\lambda^*))$. Denote $u_t = {\cal D}( V_{t} )$ for any $t \in [0, T]$. Especially, we define:
    \begin{itemize}
        \item Approximating error $\epsilon_1 := O(T^{k} s^{-k+1/2})$.
        \item Gaussian error $\epsilon_2 := O(\sqrt{d\log(d/\delta)})$.
        \item Interpolation and extrapolation error $\epsilon_3 := U d^{0.5} \sqrt{\frac{T}{\Delta t} - N} \cdot \exp(O(\frac{T}{\Delta t}s)) / \lambda^*$.
    \end{itemize}
    Then with a probability at least $1 - \delta$, we have:
    \begin{align*}
        \| \psi_t( \wt{u} ) - u_t \|_2 \leq \epsilon_1 + \epsilon_2 + \epsilon_3.
    \end{align*}
\end{lemma}

\begin{proof}{Proof sketch of Lemma~\ref{lem:hippo_error:informal}}
    This proof follows from its formal version in Lemma~\ref{lem:hippo_error}
\end{proof}

Having Lemma~\ref{lem:hippo_error:informal}, the concise bound for solving Eq.~\eqref{eq:main} could be given below:
\begin{theorem}[Informal version of Theorem~\ref{thm:inter_extra_polation}]\label{thm:inter_extra_polation:informal}
    Following Theorem~\ref{thm:uat:informal}, denote failure probability $\delta \in (0, 0.1)$ and arbitrary error $\epsilon_0 > 0$. Then with a probability at least $1 - \delta$, the network in Theorem~\ref{thm:uat:informal} satisfies Eq.~\eqref{eq:main} with $p = 2$ and
    \begin{align*}
        \epsilon = \epsilon_0 + L_0(\epsilon_1 + \epsilon_2 + \epsilon_3).
    \end{align*}
\end{theorem}

\begin{proof}[Proof sketch of Theorem~\ref{thm:inter_extra_polation:informal}]
    Please refer to Theorem~\ref{thm:inter_extra_polation} for complete proofs.
\end{proof}

{\bf Discussions.} Following the results of Lemma~\ref{lem:hippo_error:informal} and Theorem~\ref{thm:inter_extra_polation:informal}, we thus derive few insights as follows:
\begin{itemize}
    \item {\bf Optimal choice of $s$: A trade-off between $\epsilon_1$ and $\epsilon_3$. } As shown in the conditions of Lemma~\ref{lem:hippo_error:informal}, the larger value of the order of polynomials $s$ helps to decrease approximating error in the training dataset while also ruining the generalization ability.
    \ifdefined\isarxiv
    \else
    \vspace{-2mm}
    \fi
    \item {\bf Stable visual decoder. } Theorem~\ref{thm:inter_extra_polation:informal} shows a small value of $L_0$ (the stability and smoothness of visual decoder), which is important for the error of interpolation and extrapolation with an arbitrary frame rate.
    \ifdefined\isarxiv
    \else
    \vspace{-3mm}
    \fi
    \item {\bf Information. } Besides, a sub-linear factor $\sqrt{\frac{T}{\Delta t} - N}$, which stands for the obtained information about the continuous video, is vital as well for interpolation and extrapolation on data in distribution.
\end{itemize}
\ifdefined\isarxiv
\else
\vspace{-6mm}
\fi

\subsection{Timescale Robustness}\label{sub:timescale_robustness}

Following \cite{gde+20}, we demonstrate that projection onto latent patches $u_t$ is robust to timescales. Formally, the HiPPO-LegS operator is {\it timescale-equivariant}: dilating the input $u$ does not change the approximation coefficients $H_N$. At the same time, this property is working in the case of the discretized form $\wt{u}$. We emphasize that it is crucial to use flow matching to model the latent patches, where whatever the sampling method and frame rate are, it will not greatly harm VLFM's performance. We give its formal statement below.

\begin{lemma}[Proposition 3 of \cite{gde+20}, informal version of Lemma~\ref{lem:timescale_robustness}]\label{lem:timescale_robustness:informal}
    For any integer scale factor $\beta > 0$, the frames of video $\wt{V}_\tau$ is scaled to $\wt{V}_{\beta \tau}$ for each $\tau \in [\frac{T}{\Delta t}]$, it doesn’t affect the result of $H_N$.
\end{lemma}
\ifdefined\isarxiv
\else
\vspace{-3mm}
\fi
\begin{proof}
    This lemma follows from Proposition 3 in \cite{gde+20}.
\end{proof}
\ifdefined\isarxiv
\else
\vspace{-6mm}
\fi

%% file: 6_exps.tex
\begin{figure*}[!ht]
\begin{center}
\centering
    \subfloat[{\it Video caption: A green turtle swimming under the sea.}]{
    \includegraphics[width=0.95\textwidth]{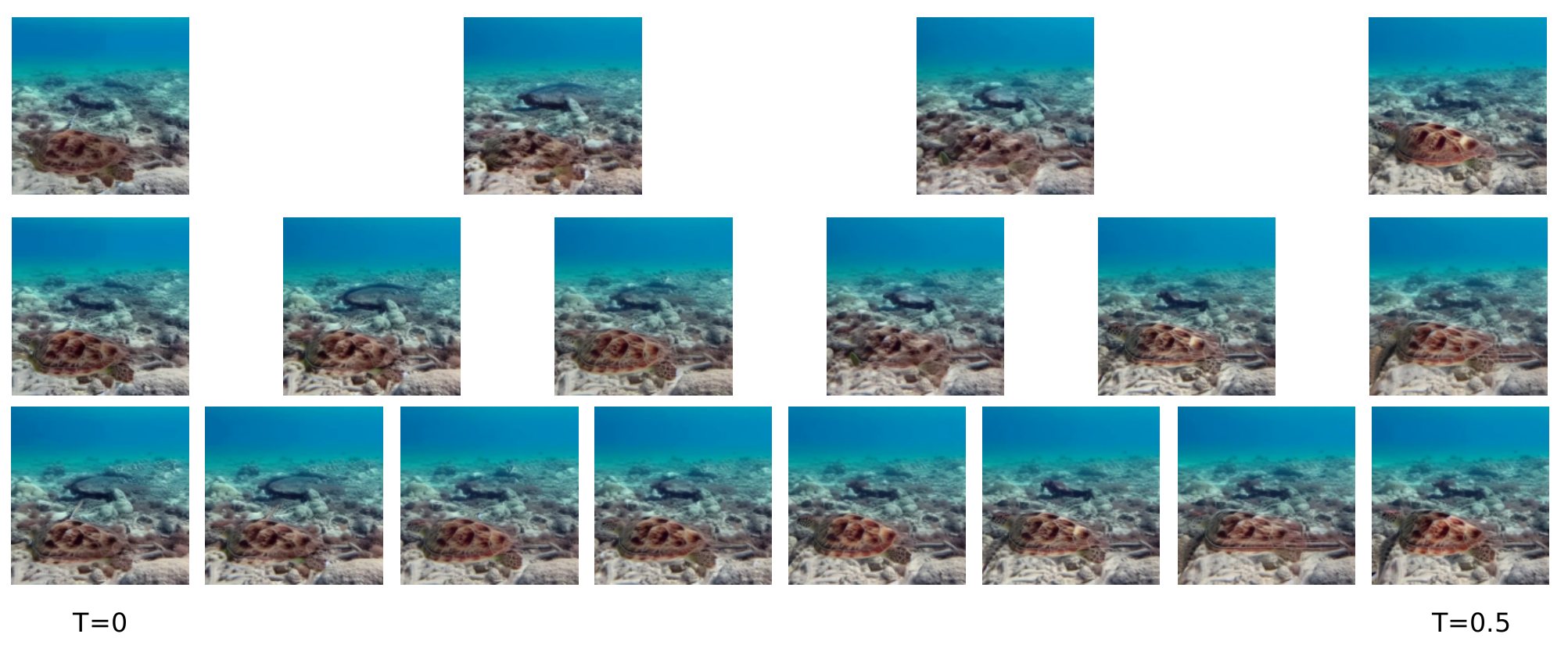}} \\
    \subfloat[{\it Video caption: Viewing countless sunflowers in a field from top.}]{
    \includegraphics[width=0.95\textwidth]{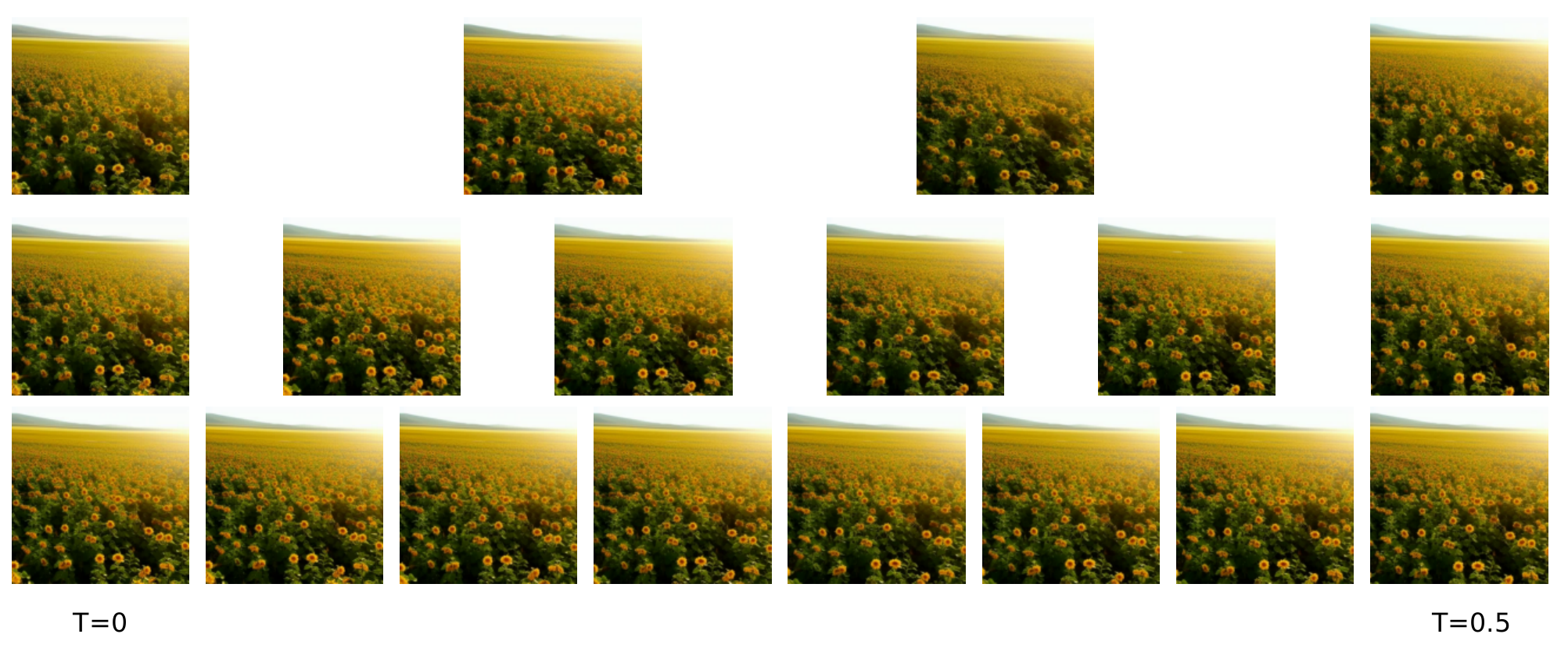}}
\end{center}
\caption{Generated videos with different frame rates $\{8, 12, 16\}$. }
\label{fig:gen}
\ifdefined\isarxiv
\else
\vspace{-3mm}
\fi
\end{figure*}

\begin{figure*}[!ht]
\begin{center}
\centering
    \subfloat{
    \includegraphics[width=0.95\textwidth]{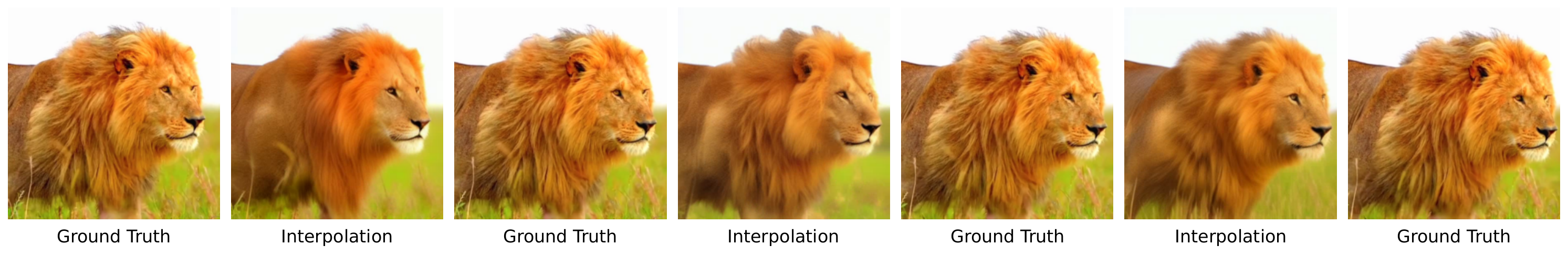}} \\
    \subfloat{
    \includegraphics[width=0.95\textwidth]{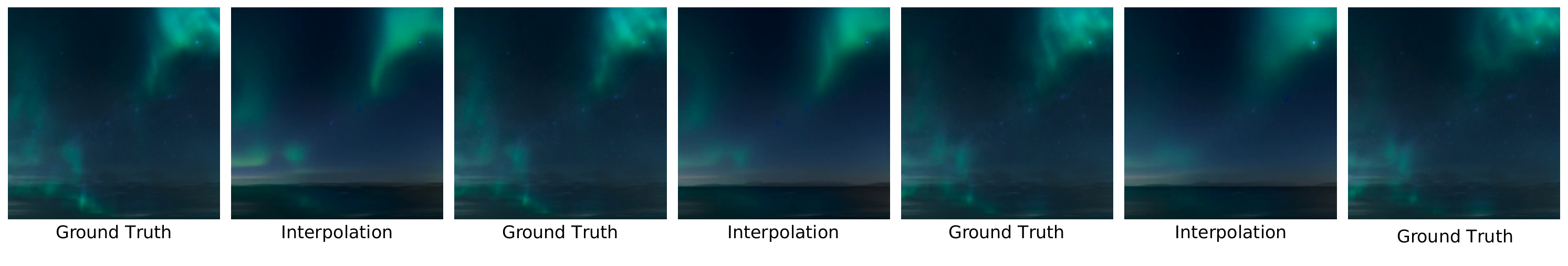}} \\
    \subfloat{
    \includegraphics[width=0.95\textwidth]{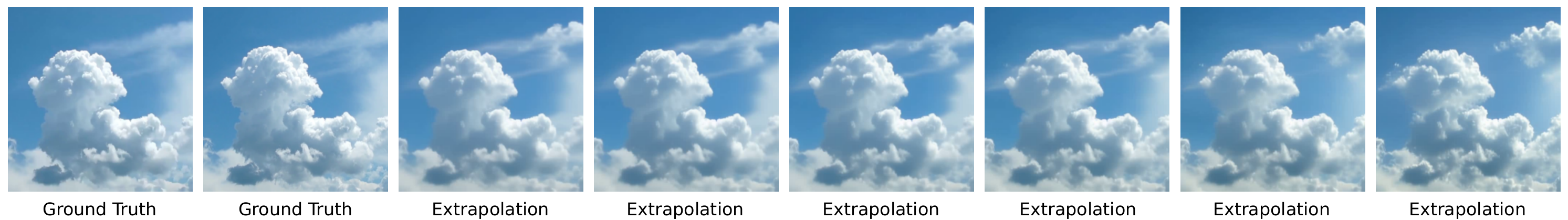}} \\
\end{center}
\caption{Interpolation and Extrapolation of VLFM.}
\label{fig:inter_extra}
\ifdefined\isarxiv
\else
\vspace{-2mm}
\fi
\end{figure*}

\section{Experiments}\label{sec:exp}

In this section, we conduct experiments to evaluate the effectiveness of our approach. We first introduce our experimental setups in Section~\ref{sub:exp_setup}. Then, we demonstrate text-to-video generation using VLFM and VLFM's capability of generating videos in arbitrary frame rate in Section~\ref{sub:exp_gen}. Furthermore, we showcase the strong performance of interpolation and extrapolation of VLFM in Section~\ref{sub:exp_inter_extra}. We also perform an ablation study to discuss the importance of the flow matching algorithm in Section~\ref{sub:exp_ablation}.

\subsection{Setup} \label{sub:exp_setup}

In our experiments, we apply Stable Diffusion v1.5 \cite{rbl+22} with DDIM scheduler \cite{sme20} as the visual decoder. Then, we use a DiT-XL-2 \cite{px23} as the backbone for the Flow Matching algorithm \cite{lcb+22,lgl22}, and the choice of hyper-parameters of $\sigma_t(\wt{u})$ is given by $\sigma_{\rm min} = 0.01$ and $\alpha = 10$. We optimize the DiT using Grams optimizer \cite{cls24}. We sample and combine 7 data resources for comprehensive training and validation of our method. They are:
OpenVid-1M \cite{nxz+24},
UCF-101 \cite{szs12},
Kinetics-400 \cite{kcs+17},
YouTube-8M \cite{akl+16},
InternVid \cite{whl+23},
MiraData \cite{jgz+24}, and
Pixabay \cite{pixabay}. 

\ifdefined\isarxiv
\else
\vspace{-4mm}
\fi

\subsection{Text-to-Video Generation with Arbitrary Frame Rate} \label{sub:exp_gen}

In this section, we recover several videos with different frame rates using VLFM with given video captions in the training dataset. We extract $T= 0.5$ for demonstrations as Figure~\ref{fig:gen}. In detail, we choose three frame rates for generation $\{8, 12, 16\}$. As shown, our VLFM performs fairly on text-to-video generation while it requires very small resource that is equivalent to training a new flow matching text-to-image video, which ensures its efficiency. Moreover, we give more results that are generated by VLFM in Appendix~\ref{sec:app:more_1} and \ref{sec:app:more_2}.
\ifdefined\isarxiv
\else
\vspace{-3mm}
\fi

\subsection{Interpolation and Extrapolation} \label{sub:exp_inter_extra}

In this section, we test the interpolation and extrapolation of VLFM. For the interpolation experiment, the model is trained with 24 FPS and evaluated to generate video with 48 FPS. For the extrapolation, the model is trained with the first video with $T = 2$ and evaluated to generate the whole video with $T = 8$. Referring the results in Figure~\ref{fig:inter_extra}, this demonstrates the strong performance of our VLFM under our mathematical guarantee of the error bound and its effectiveness.

\subsection{Ablation Study} \label{sub:exp_ablation}

In this section, we compared training VLFM with the Flow Matching algorithm and directly used DiT to predict the latent patches to showcase the importance of utilizing flow matching in our VLFM. We compare VLFM with and without flow matching by training the model with 1000 steps and compare the PSNR (peak signal-to-noise ratio) before and after training for video recovery with given captions in the training dataset. We state the results in Table~\ref{tab:ablation}. Denote ${\rm MSE}(x,y)$ as the mean squared error function, the computation of the metric PSNR is given by ($x,y \in \R^{r\times r}$):
\ifdefined\isarxiv
\else
\vspace{-3mm}
\fi
\begin{align*}
    {\rm PSNR}(x,y) := 10 \log_{10}(\frac{r^2}{{\rm MSE}(x,y)}), 
\end{align*}
\ifdefined\isarxiv
\else
\vspace{-3mm}
\fi

\begin{table}[!ht]
\ifdefined\isarxiv
\else
\vspace{-2mm}
\fi
\begin{center}
\begin{small}
\begin{sc}
\begin{tabular}{r | c c}
    \toprule
    Algorithm & Initial PSNR$\uparrow$ & Final PSNR$\uparrow$ \\
    \midrule
    Flow Matching & {\bf 57.20} & {\bf 61.18} \\
    Direct Predicting & 9.81 & 53.77 \\
    \bottomrule
\end{tabular}
\end{sc}
\end{small}
\end{center}
\caption{PSNR comparison (the greater, the better) of Flow Matching and direct generation from DiT. We boldface the better scores.}
\label{tab:ablation}
\ifdefined\isarxiv
\else
\vspace{-4mm}
\fi

\end{table}

%% file: 7_conclusion.tex
\section{Conclusion} \label{sec:conclusion}

This paper proposes {\it Video Latent Flow Matching} (VLFM) for efficient training of a time-varying flow to approximate the sequence of latent patches of the obtained video. This approach is confirmed to enjoy theoretical benefits, including 1) universal approximation theorem via applying Diffusion Transformer architecture and 2) optimal polynomial projections and timescale by introducing HiPPO-LegS. Furthermore, we provide the generalization error bound of VLFM that is trained only on the limited sub-sampled video to interpolate and extrapolate the whole ideal video. We evaluate our VLFM on Stable Diffusion v1.5 with DDIM scheduler and the DiT-XL-2 model with datasets OpenVid-1M,
UCF-101,
Kinetics-400,
YouTube-8M,
InternVid,
MiraData, and
Pixabay. The experimental results validated the potential of our approach to become a novel and efficient training form for text-to-video generation.

{\bf Limitations. } Since the motivation of this paper focuses on simply and efficiently solving the main goal, it lacks enough exploring each design and how it affects the empirical performance, providing little insights for the follow-ups. Hence, we leave these comprehensive explorations, and its more concise theoretical working mechanism behind as future works. On the other hand, although VLFM simplifies the video modeling process, it necessitates additional computational consumption concerning the combination of the visual decoder part and the flow matching part at the inference stage. We also leave such exploration to a more efficient inference method as a future direction.

%% file: 9999_impact.tex
\section*{Impact Statement}
This paper presents work whose goal is to advance the field of Machine Learning. There are many potential societal consequences of our work, none of which we feel must be specifically highlighted here.

%% file: 8_app_more_result.tex
\begin{center}
	\textbf{\LARGE Appendix }
\end{center}

In the appendix, we present more experimental text-to-video generation results in Appendix~\ref{sec:app:more_1} and more interpolation and extrapolation results in Appendix~\ref{sec:app:more_2}. Then we introduce the preliminary in Appendix~\ref{sec:app:preli}. Next, we illustrate Video Latent Flow Matching formally in Appendix~\ref{sec:app:vlfm}. In Appendix~\ref{sec:app:dit}, we demonstrate the Diffusion Transformer, and finally, in Appendix~\ref{sec:app:inter_extra}, we present the interpolation and extrapolation of VLFM.

\section{More Text-to-Video Generation Results} \label{sec:app:more_1}

\begin{figure*}[!ht]
\begin{center}
\centering
    \subfloat[{\it Video caption: Venus spinning in the space.}]{
    \includegraphics[width=0.95\textwidth]{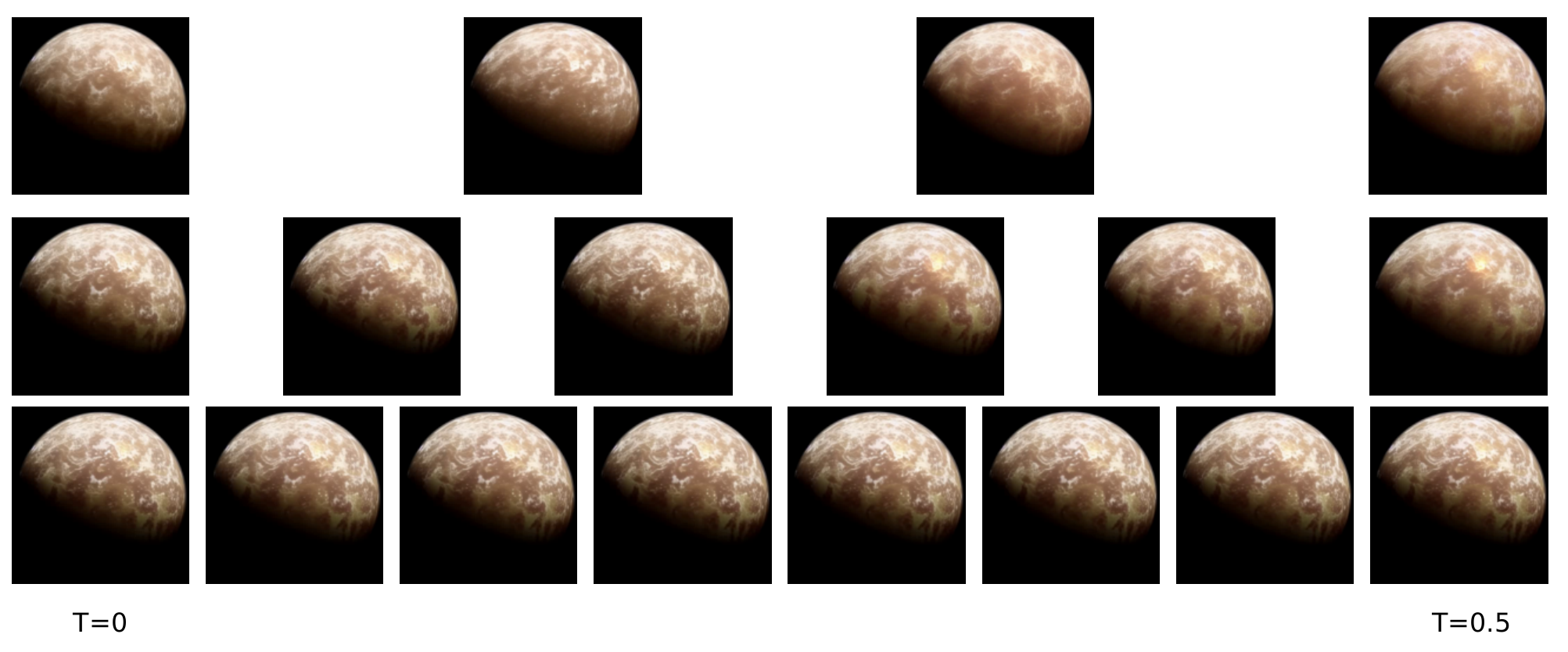}} \\
    \subfloat[{\it Video caption: Steam is coming out of a pot.}]{
    \includegraphics[width=0.95\textwidth]{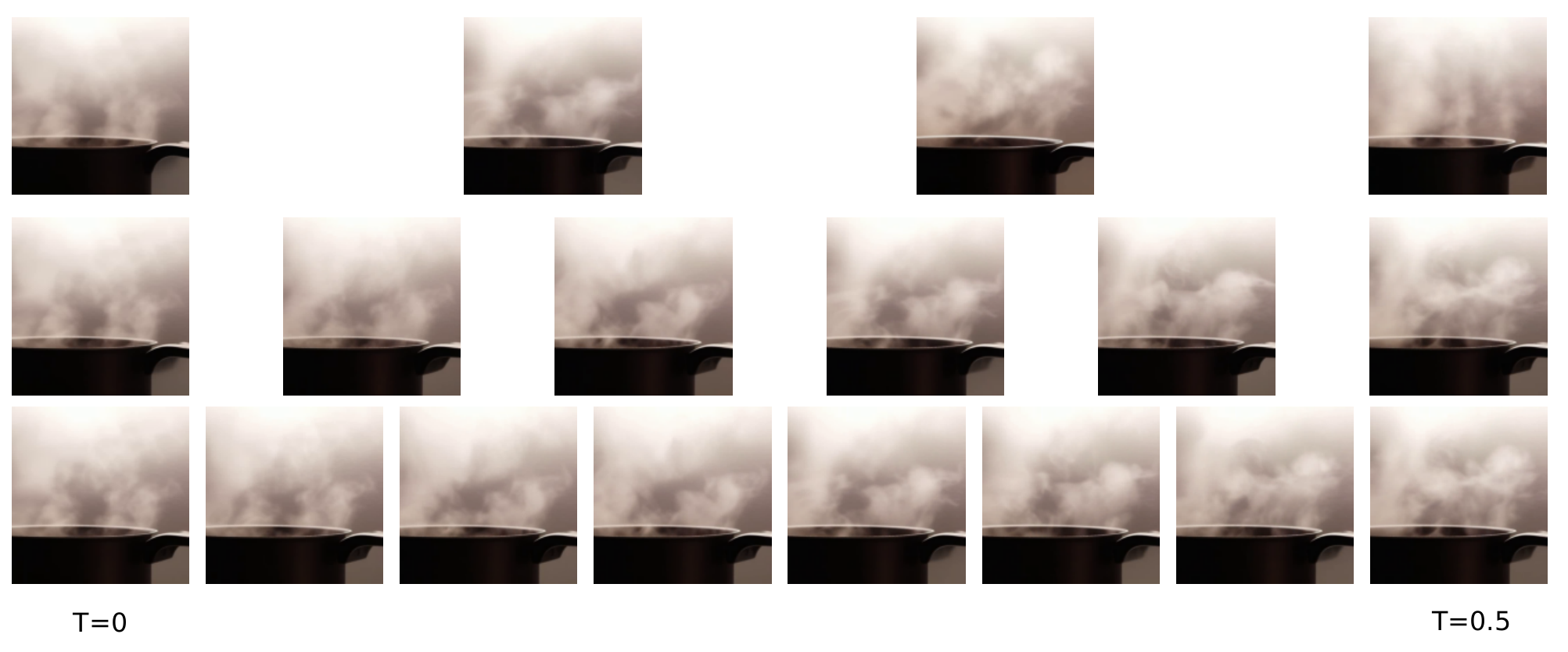}} 
\end{center}
\caption{Generated videos with different frame rates $\{8, 12, 16\}$. }
\label{fig:gen_addtional1}
\end{figure*}

\begin{figure*}[!ht]
\begin{center}
\centering
    \subfloat[{\it Video caption: Flame flickers on the candles.}]{
    \includegraphics[width=0.95\textwidth]{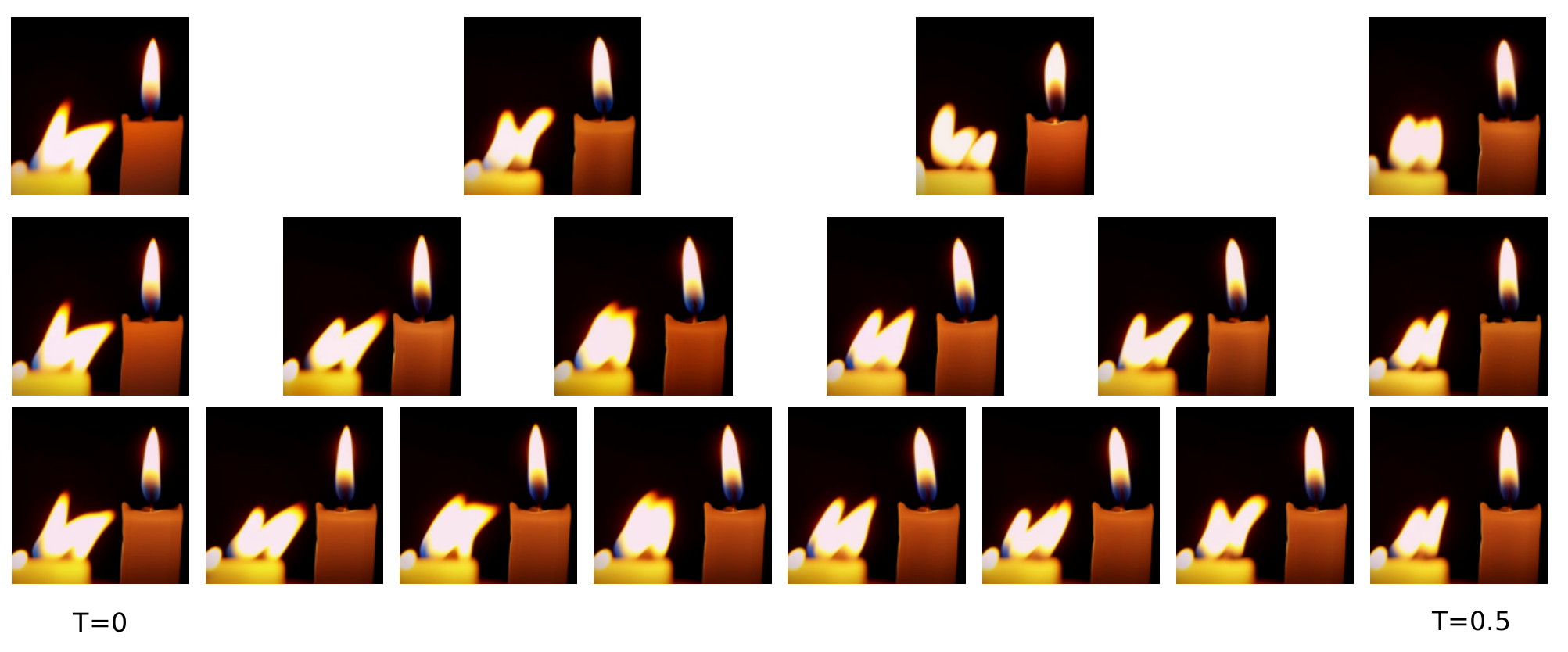}} \\
    \subfloat[{\it Video caption: A train is running through the rail road near the coast.}]{
    \includegraphics[width=0.95\textwidth]{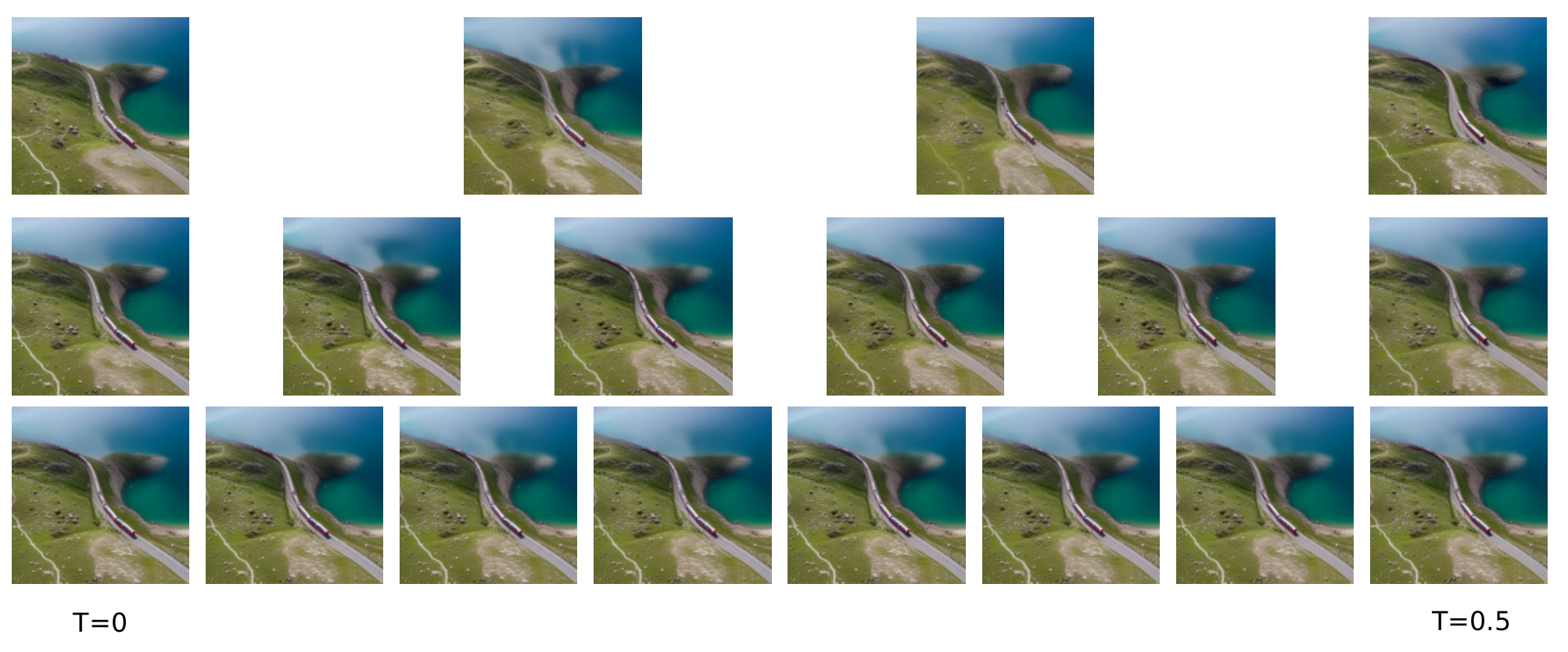}}
\end{center}
\caption{Generated videos with different frame rates $\{8, 12, 16\}$. }
\label{fig:gen_addtional2}
\end{figure*}

We give more text-to-video generation results with different frame rates to demonstrate the generative ability of our VLFM in Figure~\ref{fig:gen_addtional1} and Figure~\ref{fig:gen_addtional2}.

\section{More Interpolation and Extrapolation Results} \label{sec:app:more_2}

\begin{figure*}[!ht]
\begin{center}
\centering
    \subfloat{
    \includegraphics[width=0.95\textwidth]{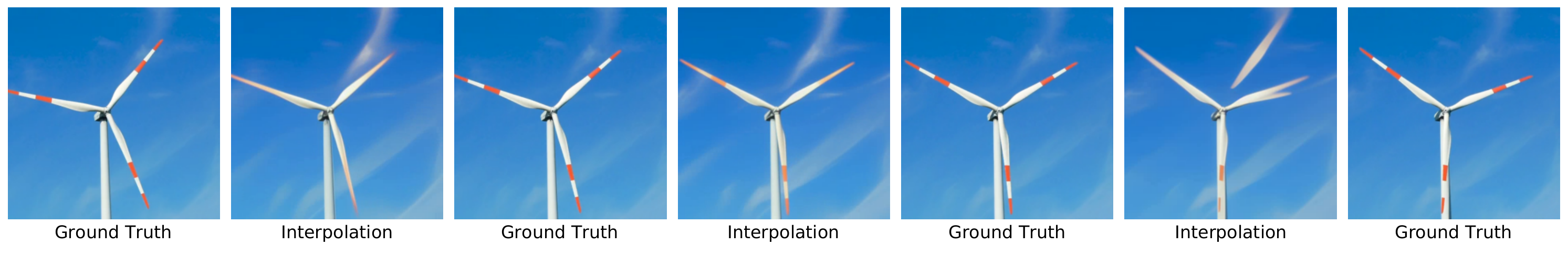}} \\
    \subfloat{
    \includegraphics[width=0.95\textwidth]{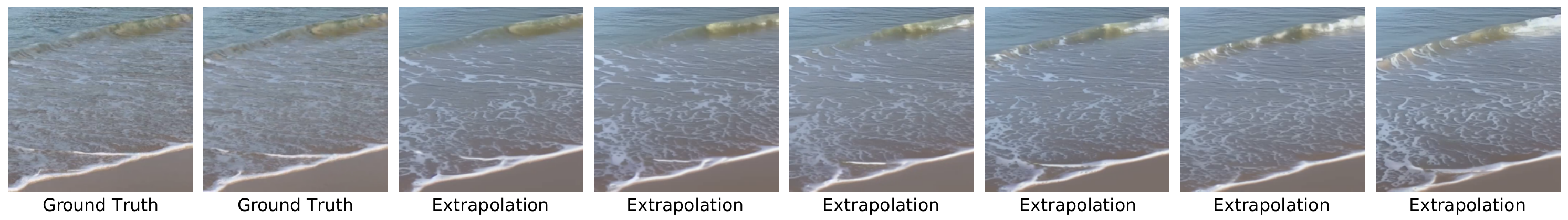}} \\
    \subfloat{
    \includegraphics[width=0.95\textwidth]{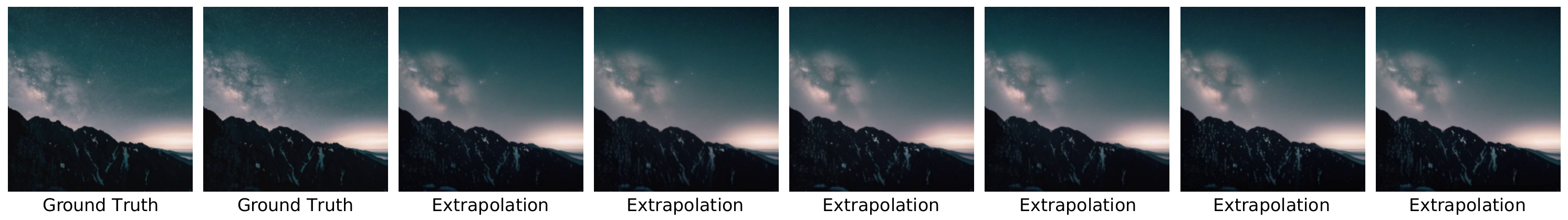}}
\end{center}
\caption{Interpolation and Extrapolation of VLFM.}
\label{fig:more_inter_extra}
\end{figure*}

We give more results of interpolation and extrapolation of VLFM in Figure~\ref{fig:more_inter_extra}.

%% file: 9_app_preli.tex
\section{Preliminary} \label{sec:app:preli}

In the preliminary section, we first introduce our notation in the appendix in Appendix~\ref{sub:app:notations}.  Then, in Appendix~\ref{sub:app:video}, we formally define the video-caption data and visual decoder. In Appendix~\ref{sub:app:latent_patches}, we define the latent patches. Appendix~\ref{sub:app:assumption} makes some assumptions which we will use later. Finally, in Appendix~\ref{sec:app:facts}, we list some basic useful facts.

\subsection{Notations} \label{sub:app:notations}

\paragraph{Notations.} We use $D$ to denote the flattened dimension of real-world images. We use $d$ to represent the dimension of latent patches. We introduce $d_0$ as the dimension of Diffusion Transformers. We utilize $V: [0, T] \rightarrow \R^D$ to denote a video with $T$ duration, where $T$ is the longest time for each video. We omit $\nabla_t a(t)$ and $a'(t)$ to denote taking differentiation to some function $a(t)$ w.r.t. time $t$. We use integer $s$ to denote the order of polynomials. The dimensional number of the text embedding vector is given by integer $\ell$.

\subsection{Video-Caption Data} \label{sub:app:video}

\begin{definition}[Video-caption data pairs and their distribution]\label{def:V_c}
    We define a video caption distribution $(V, c) \sim {\cal V}_c$. Here, $V: [0, T] \rightarrow \R^D$ is considered as a function and $c \in \R^\ell$ is the corresponding text embedding vector.
\end{definition}

\begin{definition}\label{def:wt_V}
    Given a video caption distribution ${\cal V}_c$ as Definition~\ref{def:V_c}. We denote $\Delta t$ as the minimal time unit of measurement in the real world (Planck time). For any $(V, c) \sim {\cal V}_c$, we define the discretized form of $V: [0, T] \rightarrow \R^D$, which is $\wt{V} \in \R^{\frac{T}{\Delta t} \times D}$, and its $\tau$-th row $ \forall \tau \in [\frac{T}{\Delta t}]$ is given by:
    \begin{align*}
        \wt{V}_\tau := V_{\Delta t \cdot \tau} \in \R^D.
    \end{align*}
\end{definition}

\begin{definition}[Obtained data in real-world cases]\label{def:Phi}
    If the following conditions hold:
    \begin{itemize}
        \item Given a video caption distribution ${\cal V}_c$ as Definition~\ref{def:V_c}.
        \item For any $(V, c) \sim {\cal V}_c$, we define the discretized form of video $\wt{V}$ as Definition~\ref{def:wt_V}.
    \end{itemize}
    We define an observation matrix $\Phi: \{0, 1\}^{N \times \frac{T}{\Delta t}}$. The obtained data in real-world cases then is denoted as $\Phi \wt{V} \in \R^{N \times D}$.
\end{definition}

\begin{definition}[Bijective Visual Decoder]\label{def:visual_decoder}
    We define the visual decoder ${\cal D}: \R^d \rightarrow \R^D$ satisfies that:
    \begin{itemize}
        \item For any flattened image $V \in \R^D$, there is a unique $u \in \R^d$ such that ${\cal D}(u) = V$.
    \end{itemize}
    Then we say ${\cal D}$ is bijective. Denote the reverse function of ${\cal D}$ as ${\cal D}^{-1}: \R^D \rightarrow \R^d$.
\end{definition}

\subsection{Latent Patches Data} \label{sub:app:latent_patches}

\begin{definition}\label{def:u}
    If the following conditions hold:
    \begin{itemize}
        \item Given a video caption distribution ${\cal V}_c$ as Definition~\ref{def:V_c}.
        \item For any $(V, c) \sim {\cal V}_c$, we define the discretized form of video $\wt{V}$ as Definition~\ref{def:wt_V}.
        \item Let the observation matrix $\Phi: \{0, 1\}^{N \times \frac{T}{\Delta t}}$ be defined as Definition~\ref{def:Phi}.
        \item Let the visual decoder function $D: \R^d \rightarrow \R^D$ be defined as Definition~\ref{def:visual_decoder}.
    \end{itemize}
    We define the ideal version (without observation matrix) of the sequence of latent patches $u \in \R^{\frac{T}{\Delta t} \times d}$, and its $\tau$-th $ \forall \tau \in [\frac{T}{\Delta t}]$ row is defined as follows:
    \begin{align*}
        u_\tau := {\cal D}^{-1}( \wt{V}_\tau ).
    \end{align*}
\end{definition}

\begin{definition}\label{def:wt_u}
    If the following conditions hold:
    \begin{itemize}
        \item Given a video caption distribution ${\cal V}_c$ as Definition~\ref{def:V_c}.
        \item For any $(V, c) \sim {\cal V}_c$, we define the discretized form of video as Definition~\ref{def:wt_V}.
        \item Let the observation matrix $\Phi: \{0, 1\}^{N \times \frac{T}{\Delta t}}$ be defined as Definition~\ref{def:Phi}.
        \item Let the visual decoder function $D: \R^d \rightarrow \R^D$ be defined as Definition~\ref{def:visual_decoder}.
    \end{itemize}
    We define the real-world version (with observation matrix) of the sequence of latent patches $\wt{u} \in \R^{\frac{T}{\Delta t} \times d}$, and its $\tau$-th $ \forall \tau \in [N]$ row is defined as follows:
    \begin{align*}
        \wt{u}_\tau := {\cal D}^{-1}\Big( (\Phi V)_\tau \Big).
    \end{align*}
\end{definition}

\subsection{Assumptions} \label{sub:app:assumption}

\begin{assumption}\label{ass:k}
    If the following conditions hold:
    \begin{itemize}
        \item Given a video caption distribution ${\cal V}_c$ as Definition~\ref{def:V_c}.
        \item For any $(V, c) \sim {\cal V}_c$, we define the discretized form of video as Definition~\ref{def:wt_V}.
        \item Let the observation matrix $\Phi: \{0, 1\}^{N \times \frac{T}{\Delta t}}$ be defined as Definition~\ref{def:Phi}.
        \item Let the visual decoder function $D: \R^d \rightarrow \R^D$ be defined as Definition~\ref{def:visual_decoder}.
        \item Let the ideal version of the sequence of latent patches $u \in \R^{\frac{T}{\Delta t} \times d}$ be defined as Definition~\ref{def:u}.
    \end{itemize}
    We assume $u_\tau$ is $k$-differentiable, there exists:
    \begin{align*}
        u_{\tau}^{(i)} = \lim_{\Delta t \rightarrow 0} \frac{u_{\tau+1}^{(i-1)} - u_{\tau}^{(i-1)}}{ \Delta t }, \forall i \in [k], \tau \in [\frac{T}{\Delta t}],
    \end{align*}
    where, we use $u_\tau^{(i)}$ to denote the $i$-th derivation of $u$.
\end{assumption}

\begin{assumption}\label{ass:L_0}
    If the following conditions hold:
    \begin{itemize}
        \item Let the visual decoder function $D: \R^d \rightarrow \R^D$ be defined as Definition~\ref{def:visual_decoder}.
    \end{itemize}
    We assume the visual decoder function ${\cal D}$ is $L_0$-smooth for constant $L_0 > 0$, such that:
    \begin{align*}
        \| {\cal D}(x) - {\cal D}(y) \|_2 \leq L_0 \| x - y \|_2, \forall x, y \in \R^d.
    \end{align*}
\end{assumption}

\begin{assumption}\label{ass:U}
    If the following conditions hold:
    \begin{itemize}
        \item Given a video caption distribution ${\cal V}_c$ as Definition~\ref{def:V_c}.
        \item For any $(V, c) \sim {\cal V}_c$, we define the discretized form of video as Definition~\ref{def:wt_V}.
        \item Let the observation matrix $\Phi: \{0, 1\}^{N \times \frac{T}{\Delta t}}$ be defined as Definition~\ref{def:Phi}.
        \item Let the visual decoder function $D: \R^d \rightarrow \R^D$ be defined as Definition~\ref{def:visual_decoder}.
        \item Let the ideal version of the sequence of latent patches $u \in \R^{\frac{T}{\Delta t} \times d}$ be defined as Definition~\ref{def:u}.
    \end{itemize}
    We assume each entry in latent patches $u$ is bounded by a constant $U > 0$.
\end{assumption}

\begin{assumption}\label{ass:M}
    If the following conditions hold:
    \begin{itemize}
        \item Given a video caption distribution ${\cal V}_c$ as Definition~\ref{def:V_c}.
        \item For any $(V, c) \sim {\cal V}_c$
    \end{itemize}
    For any $(V, c) \sim {\cal V}_c$, we assume there exists a function ${\cal M}: [0, T] \times \R^\ell \rightarrow \R^D$ satisfies $V_t = {\cal M}_t(c)$. 
\end{assumption}

\subsection{Basic Facts} \label{sec:app:facts}

\begin{fact}\label{fac:gaussian_tail}
    For a variable $x \sim \mathcal{N}(0, \sigma^2)$, then with probability at least $1 - \delta$, we have:
    \begin{align*}
        |x| \leq C \sigma \sqrt{\log(1/\delta)}
    \end{align*}
\end{fact}

\begin{fact}\label{fac:infity_norm_pesdueo_inverse}
    For a PD matrix $A \in \R^{d_1 \times d_2}$ with a positive minimum eigenvalue $\lambda_{\min}(A) > 0$, the infinite norm of its pesdueo-inverse matrix $A^\dag$ is given by:
    \begin{align*}
        \| A^\dagger \|_\infty \leq \frac{1}{\lambda_{\min}(A)}.
    \end{align*}
\end{fact}

\begin{fact}\label{fac:pesdueo_inverse_diff}
    For two matrices $A , B \in\R^{d_1 \times d_2}$, we have:
    \begin{align*}
        \| A^\dagger - B^\dagger \|_2 \leq \frac{\| A^\dagger \|_2^2 \| A - B\|_2 }{1 - \| A^\dagger \|_2 \cdot 
        \| A - B \|_2}
    \end{align*}
\end{fact}

%% file: 10_app_vlfm.tex
\section{Video Latent Flow Matching}
\label{sec:app:vlfm}

This section, we first introduce the HiPPO Framework and LegS in Appendix~\ref{sub:app:hippo}. Then, we formally define the video latent flow in Appendix~\ref{sub:app:vlf}. Last, we introduce the training objective of VLFM in Appendix~\ref{sub:app:train_obj}.

\subsection{HiPPO Framework and LegS} \label{sub:app:hippo}

\begin{definition}\label{def:A}
    We define matrix $A \in \R^{s \times s}$ where its $(i_1, i_2)$-th entry $\forall i_1, i_2 \in [s]$ is given by:
    \begin{align*}
        A_{i_1, i_2} & ~ = \begin{cases}
        \sqrt{(2i_1 + 1)(2i_2 + 1)}, & \text{if $i_1 > i_2$} \\
        i_1 + 1, & \text{if $i_1 = i_2$} \\
        0, & \text{if $i_1 < i_2$}
    \end{cases}.
    \end{align*}
\end{definition}

\begin{definition}\label{def:B}
    We define matrix $B \in \R^{s \times 1}$ where its $i_1$-th entry $\forall i_1 \in [s]$ is given by:
    \begin{align*}
        B_{i_1} & ~ = \sqrt{2i_1 + 1}.
    \end{align*}
\end{definition}

\begin{definition}\label{def:H}
    If the following conditions hold:
    \begin{itemize}
        \item Let matrix $A \in \R^{s \times s}$ be defined as Definition~\ref{def:A}.
        \item Let matrix $B \in \R^{s \times 1}$ be defined as Definition~\ref{def:B}.
    \end{itemize}
    We initialize a matrix $H_0 = {\bf 0}_{d \times s}$. Then we define:
    \begin{align*}
        H_{\tau} := H_{\tau-1}( I_s - \frac{1}{\tau} A )^\top + \frac{1}{\tau} \wt{u}_\tau B^\top, \forall \tau \in [N].
    \end{align*}
\end{definition}

\begin{definition}\label{def:g_t}
    We define $g(t) := [\sqrt{\frac{1}{2}} P_0(t), \sqrt{\frac{3}{2}} P_1(t), \cdots, \sqrt{\frac{2s-1}{2}} P_{s-1}(t)]^\top $ $\in \R^{s} $, where$ P_i(t), \forall i \in [s]$ is some polynomials. Especially, $g(t)$ satisfies:
    \begin{itemize}
        \item Define $G := \begin{bmatrix}
            g(\Delta t)^\top \\
            g(2 \Delta t)^\top \\
            \vdots \\
            g(T)^\top
        \end{bmatrix}$, $\lambda_{\min} (G) > 0$. Here, $\lambda_{\min}$ is the function that outputs the minimal eigenvalue of the input matrix.
        \item $|G_{\tau, i}| \leq \exp(O(\frac{T}{\Delta t}s))$ for any $\tau \in [\frac{T}{ \Delta t}], i \in [s]$.
    \end{itemize}
\end{definition}

\subsection{Video Latent Flow} \label{sub:app:vlf}

\begin{definition}\label{def:mu}
    If the following conditions hold:
    \begin{itemize}
        \item Given a video caption distribution ${\cal V}_c$ as Definition~\ref{def:V_c}.
        \item For any $(V, c) \sim {\cal V}_c$, we define the discretized form of video as Definition~\ref{def:wt_V}.
        \item Let the observation matrix $\Phi: \{0, 1\}^{N \times \frac{T}{\Delta t}}$ be defined as Definition~\ref{def:Phi}.
        \item Let the visual decoder function $D: \R^d \rightarrow \R^D$ be defined as Definition~\ref{def:visual_decoder}.
        \item Let the ideal version of the sequence of latent patches $u \in \R^{\frac{T}{\Delta t} \times d}$ be defined as Definition~\ref{def:u}.
        \item Let the real-world version of the sequence of latent patches $\wt{u} \in \R^{N \times d}$ be defined as Definition~\ref{def:wt_u}.
        \item Let $H_N \in \R^{d \times s}$ be defined as Definition~\ref{def:H}.
        \item Let the function of polynomials $g(t)$ be defined as Definition~\ref{def:g_t}.
    \end{itemize}
    We define the time-dependent mean of Gaussian distribution as follows:
    \begin{align*}
        \mu_t(\wt{u}) := H_N g(t) \in \R^d
    \end{align*}
\end{definition}

\begin{definition}\label{def:sigma}
    If the following conditions hold:
    \begin{itemize}
        \item Given a video caption distribution ${\cal V}_c$ as Definition~\ref{def:V_c}.
        \item For any $(V, c) \sim {\cal V}_c$, we define the discretized form of video as Definition~\ref{def:wt_V}.
        \item Let the observation matrix $\Phi: \{0, 1\}^{N \times \frac{T}{\Delta t}}$ be defined as Definition~\ref{def:Phi}.
        \item Let the visual decoder function $D: \R^d \rightarrow \R^D$ be defined as Definition~\ref{def:visual_decoder}.
        \item Let the ideal version of the sequence of latent patches $u \in \R^{\frac{T}{\Delta t} \times d}$ be defined as Definition~\ref{def:u}.
        \item Let the real-world version of the sequence of latent patches $\wt{u} \in \R^{N \times d}$ be defined as Definition~\ref{def:wt_u}.
        \item Let $H_N \in \R^{d \times s}$ be defined as Definition~\ref{def:H}.
        \item Let the function of polynomials $g(t)$ be defined as Definition~\ref{def:g_t}.
        \item Denote $\sigma_{\min} > 0$.
        \item Given a hyper-parameter $\alpha > 0$.
    \end{itemize}
    We define  the time-dependent standard deviation as follows:
    \begin{align*}
        \sigma_t(\wt{u}) := (1 - \sigma_{\min}) \cdot [\sin^2( \pi \frac{N}{T} t ) +  \exp(-\alpha t) ] + \sigma_{\min} \in \R_{\ge 0}.
    \end{align*}
\end{definition}

\begin{lemma}\label{lem:bound_diff_sigma}
    If the following conditions hold:
    \begin{itemize}
        \item Given a video caption distribution ${\cal V}_c$ as Definition~\ref{def:V_c}.
        \item For any $(V, c) \sim {\cal V}_c$, we define the discretized form of video as Definition~\ref{def:wt_V}.
        \item Let the observation matrix $\Phi: \{0, 1\}^{N \times \frac{T}{\Delta t}}$ be defined as Definition~\ref{def:Phi}.
        \item Let the visual decoder function $D: \R^d \rightarrow \R^D$ be defined as Definition~\ref{def:visual_decoder}.
        \item Let the ideal version of the sequence of latent patches $u \in \R^{\frac{T}{\Delta t} \times d}$ be defined as Definition~\ref{def:u}.
        \item Let the real-world version of the sequence of latent patches $\wt{u} \in \R^{N \times d}$ be defined as Definition~\ref{def:wt_u}.
        \item Let $H_N \in \R^{d \times s}$ be defined as Definition~\ref{def:H}.
        \item Let the function of polynomials $g(t)$ be defined as Definition~\ref{def:g_t}.
        \item Let the time-dependent mean of Gaussian distribution $\mu_t(\wt{u})$ be defined as Definition~\ref{def:mu}.
        \item Let the time-dependent standard deviation $\sigma_t(\wt{u})$ be defined as Definition~\ref{def:sigma}.
        \item Denote $\sigma_{\min} > 0$.
        \item Given a hyper-parameter $\alpha > 0$.
    \end{itemize}
    Then for any $\alpha >0$, we have:
    \begin{align*}
        | \frac{\sigma_t'(\wt{u})}{\sigma_t(\wt{u})} | \leq \frac{1 - \sigma_{\min}}{\sigma_{\min}}.
    \end{align*}
\end{lemma}

\begin{proof}
    This result can be obtained following very simple algebras.
\end{proof}

\begin{definition}\label{def:psi}
    If the following conditions hold:
    \begin{itemize}
        \item Given a video caption distribution ${\cal V}_c$ as Definition~\ref{def:V_c}.
        \item For any $(V, c) \sim {\cal V}_c$, we define the discretized form of video as Definition~\ref{def:wt_V}.
        \item Let the observation matrix $\Phi: \{0, 1\}^{N \times \frac{T}{\Delta t}}$ be defined as Definition~\ref{def:Phi}.
        \item Let the visual decoder function $D: \R^d \rightarrow \R^D$ be defined as Definition~\ref{def:visual_decoder}.
        \item Let the ideal version of the sequence of latent patches $u \in \R^{\frac{T}{\Delta t} \times d}$ be defined as Definition~\ref{def:u}.
        \item Let the real-world version of the sequence of latent patches $\wt{u} \in \R^{N \times d}$ be defined as Definition~\ref{def:wt_u}.
        \item Let $H_N \in \R^{d \times s}$ be defined as Definition~\ref{def:H}.
        \item Let the function of polynomials $g(t)$ be defined as Definition~\ref{def:g_t}.
        \item Let the time-dependent mean of Gaussian distribution $\mu_t(\wt{u})$ be defined as Definition~\ref{def:mu}.
        \item Let the time-dependent standard deviation $\sigma_t(\wt{u})$ be defined as Definition~\ref{def:sigma}.
        \item Denote $\sigma_{\min} > 0$.
        \item Sample $z \sim \mathcal{N}(0, I_d)$.
    \end{itemize}
    We define the Video Latent Flow:
    \begin{align*}
        \psi_t(\wt{u}) := \sigma_t(\wt{u}) \cdot z + \mu_t(\wt{u}) \in \R^d.
    \end{align*}
\end{definition}

\subsection{Training Objective} \label{sub:app:train_obj}

\begin{definition}\label{def:L}
    If the following conditions hold:
    \begin{itemize}
        \item Given a video caption distribution ${\cal V}_c$ as Definition~\ref{def:V_c}.
        \item For any $(V, c) \sim {\cal V}_c$, we define the discretized form of video as Definition~\ref{def:wt_V}.
        \item Let the observation matrix $\Phi: \{0, 1\}^{N \times \frac{T}{\Delta t}}$ be defined as Definition~\ref{def:Phi}.
        \item Let the visual decoder function $D: \R^d \rightarrow \R^D$ be defined as Definition~\ref{def:visual_decoder}.
        \item Let the ideal version of the sequence of latent patches $u \in \R^{\frac{T}{\Delta t} \times d}$ be defined as Definition~\ref{def:u}.
        \item Let the real-world version of the sequence of latent patches $\wt{u} \in \R^{N \times d}$ be defined as Definition~\ref{def:wt_u}.
        \item Let $H_N \in \R^{d \times s}$ be defined as Definition~\ref{def:H}.
        \item Let the function of polynomials $g(t)$ be defined as Definition~\ref{def:g_t}.
        \item Let the time-dependent mean of Gaussian distribution $\mu_t(\wt{u})$ be defined as Definition~\ref{def:mu}.
        \item Let the time-dependent standard deviation $\sigma_t(\wt{u})$ be defined as Definition~\ref{def:sigma}.
        \item Denote $\sigma_{\min} > 0$.
        \item Sample $z \sim \mathcal{N}(0, I_d)$.
        \item Define a model function $F_\theta: \R^d \times \R^\ell \times [0, T] \rightarrow \R^d$ with parameters $\theta$.
    \end{itemize}
    We define the training objective of Video Latent Flow Matching as follows:
    \begin{align*}
        {\cal L}(\theta) := \E_{z \sim \mathcal{N}(0, I_d), t \sim {\sf Uniform}[0, T], (V, c) \sim {\cal V}_c}[\| F_\theta( \psi_t(\wt{u}), c, t ) - \frac{\d }{\d t} \psi_t(\wt{u}) \|_2^2].
    \end{align*}
\end{definition}

\begin{theorem}\label{thm:close_form}
    If the following conditions hold:
    \begin{itemize}
        \item Given a video caption distribution ${\cal V}_c$ as Definition~\ref{def:V_c}.
        \item For any $(V, c) \sim {\cal V}_c$, we define the discretized form of video as Definition~\ref{def:wt_V}.
        \item Let the observation matrix $\Phi: \{0, 1\}^{N \times \frac{T}{\Delta t}}$ be defined as Definition~\ref{def:Phi}.
        \item Let the visual decoder function $D: \R^d \rightarrow \R^D$ be defined as Definition~\ref{def:visual_decoder}.
        \item Let the ideal version of the sequence of latent patches $u \in \R^{\frac{T}{\Delta t} \times d}$ be defined as Definition~\ref{def:u}.
        \item Let the real-world version of the sequence of latent patches $\wt{u} \in \R^{N \times d}$ be defined as Definition~\ref{def:wt_u}.
        \item Let $H_N \in \R^{d \times s}$ be defined as Definition~\ref{def:H}.
        \item Let the function of polynomials $g(t)$ be defined as Definition~\ref{def:g_t}.
        \item Let the time-dependent mean of Gaussian distribution $\mu_t(\wt{u})$ be defined as Definition~\ref{def:mu}.
        \item Let the time-dependent standard deviation $\sigma_t(\wt{u})$ be defined as Definition~\ref{def:sigma}.
        \item Denote $\sigma_{\min} > 0$.
        \item Sample $z \sim \mathcal{N}(0, I_d)$.
        \item Define a model function $F_\theta: \R^d \times \R^\ell \times [0, T] \rightarrow \R^d$ with parameters $\theta$.
        \item Let the training objective ${\cal L}(\theta)$ be defined as Definition~\ref{def:L}.
    \end{itemize}
    Then the minimum solution for function $F_\theta$ that takes $z \sim N(0, I_d)$ and $t \sim {\sf Uniform}[0, T]$ is:
    \begin{align*}
        F_\theta(z, c, t) = \frac{\sigma_t'(\wt{u})}{\sigma_t(\wt{u})} (z - \mu_t(\wt{u})) + \mu_t'(\wt{u}).
    \end{align*}
\end{theorem}

\begin{proof}
    This proof follows from Theorem 3 in \cite{lcb+22}.
\end{proof}

%% file: 11_app_dit.tex
\section{Diffusion Transformer} \label{sec:app:dit}

In this section, we first define the Diffusion Transformer in Appendix~\ref{sub:app:def}. Moreover, we introduce the Approximation via DiT in Appendix~\ref{sub:app:approx_dit}.

\subsection{Definitions} \label{sub:app:def}

\begin{definition}[Multi-head self-attention]\label{def:attn}
    Given $h$-heads query, key, value and output projection weights $\{(W_Q^i, W_K^i, W_V^i, W_O^i)\}_{i=1}^h \subset \R^{d_0 \times 4m}$ with each weight is a $d_0 \times m$ shape matrix, for an input matrix $X \in \R^{n \times d_0}$, we define a multi-head self-attention computation as follows:
    \begin{align*}
        {\sf Attn}(X) := \sum_{i=1}^h {\sf Softmax}( X W_Q^i {W_K^i}^\top X^\top ) \cdot X W_V^i {W_O^i}^\top + X \in \R^{n \times d_0}.
    \end{align*}
\end{definition}

\begin{definition}[Feed-forward]\label{def:feed_forward}
    Given two projection weights $W_1, W_2 \in \R^{d_0 \times r}$ and two bias vectors $b_1 \in \R^r$ and $b_2 \in \R^{d_0}$, for an input matrix $X \in \R^{n \times d_0}$, we define a feed-forward computation as follows:
    \begin{align*}
        {\sf FF}(X) := \phi(X W_1 + {\bf 1}_n b_1^\top) \cdot W_2^\top + {\bf 1}_n b_2^\top + X \in \R^{n \times d_0}.
    \end{align*}
    Here, $\phi$ is an activation function and usually be considered as ReLU.
\end{definition}

\begin{definition}[Transformer block]\label{def:transformer_tf}
    Given a set of model weights $\theta^{h, m, r} = \{ \{(W_Q^i, W_K^i, W_V^i, W_O^i)\}_{i=1}^h,$ $ W_1, W_2, b_1, b_2 \}$, the computation of a transformer block is given by the combination of multi-head self-attention computation (Definition~\ref{def:attn}) and feed-forward computation (Definition~\ref{def:feed_forward}). Formally, for an input matrix $X \in \R^{n \times d_0}$, we define:
    \begin{align*}
        {\sf TF}_{\theta^{h, m, r}}(X) := {\sf FF} \circ {\sf Attn}(X) \in \R^{n \times d_0}
    \end{align*}
\end{definition}

\begin{definition}[Reshape Layer]\label{def:R}
    We define the reshape network $R: \R^d \rightarrow \R^{n \times d_0}$.
\end{definition}

\begin{definition}[Complete transformer network]\label{def:model}
    We consider a transformer network as a composition of a transformer block (Definition~\ref{def:transformer_tf}) with model weight $\theta^{h, m, r}$, which is:
    \begin{align*}
        & ~ {\cal T}^{h, m, r} \\
        := & ~ \{ {\cal F}: \R^{n \times d_0} \rightarrow \R^{n \times d_0}~\\
        & ~ |~\text{${\cal F}$ is a composition of Transformer blocks ${\sf TF}_{\theta^{h, m, r}}$’s with positional embedding $E \in \R^{n \times d_0}$}\}
    \end{align*}
    We especially say $\theta^{h, m, r}$ is the model weight that contains $h$ heads, $m$ hidden size for attention and $r$ hidden size for feed-forward. See Example~\ref{exp:cal_F} for further explanation of the sequence-to-sequence mapping ${\cal F}$.
\end{definition}

\begin{example}\label{exp:cal_F}
    We here give an example for the sequence-to-sequence mapping ${\cal F}$ in Definition~\ref{def:model}: Denote $L$ as the number of layers in some transformer network. For an input matrix $X \in \R^{n \times d}$, we use $E \in \R^{n \times d}$ to denote the positional encoding, we then define:
    \begin{align*}
        {\cal F}(X) := {\sf TF}^L \circ {\sf TF}^{L-1} \circ \cdots \circ {\sf TF}^2 \circ {\sf TF}^1(X + E)
    \end{align*}
\end{example}

\subsection{Approximation via DiT} \label{sub:app:approx_dit}

\begin{theorem}\label{thm:uat}
    If the following conditions hold:
    \begin{itemize}
        \item Given a video caption distribution ${\cal V}_c$ as Definition~\ref{def:V_c}.
        \item For any $(V, c) \sim {\cal V}_c$, we define the discretized form of video as Definition~\ref{def:wt_V}.
        \item Let the observation matrix $\Phi: \{0, 1\}^{N \times \frac{T}{\Delta t}}$ be defined as Definition~\ref{def:Phi}.
        \item Let the visual decoder function $D: \R^d \rightarrow \R^D$ be defined as Definition~\ref{def:visual_decoder}.
        \item Let the ideal version of the sequence of latent patches $u \in \R^{\frac{T}{\Delta t} \times d}$ be defined as Definition~\ref{def:u}.
        \item Let the real-world version of the sequence of latent patches $\wt{u} \in \R^{N \times d}$ be defined as Definition~\ref{def:wt_u}.
        \item Let $H_N \in \R^{d \times s}$ be defined as Definition~\ref{def:H}.
        \item Let the function of polynomials $g(t)$ be defined as Definition~\ref{def:g_t}.
        \item Let the time-dependent mean of Gaussian distribution $\mu_t(\wt{u})$ be defined as Definition~\ref{def:mu}.
        \item Let the time-dependent standard deviation $\sigma_t(\wt{u})$ be defined as Definition~\ref{def:sigma}.
        \item Denote $\sigma_{\min} > 0$.
        \item Sample $z \sim \mathcal{N}(0, I_d)$.
        \item Define a model function $F_\theta: \R^d \times \R^\ell \times [0, T] \rightarrow \R^d$ with parameters $\theta$.
        \item Let the training objective ${\cal L}(\theta)$ be defined as Definition~\ref{def:L}.
    \end{itemize}
    Then there exists a transformer network $f_{\cal T} \in {\cal T}_{P}^{2, 1, 4}$ defining function $F_\theta(z, c, t) := f_{\cal T}( R([z^\top, c^\top, t]^\top) )$ with parameters $\theta$ that satisfies ${\cal L}(\theta) \leq \epsilon$ for any error $\epsilon > 0$.
\end{theorem}

\begin{proof}
    Following Assumption~\ref{ass:M}, we first denote $\wt{V}_{\tau} = \wt{{\cal M}}_\tau(c)$ for any $\tau \in [\frac{T}{\Delta t}]$ to discretize function ${\cal M}$. Then we have:
    \begin{align}\label{eq:wt_u_func}
        \wt{u}_{\tau} = {\cal D}^{-1}\Big( (\Phi \wt{{\cal M}}(c))_\tau \Big).
    \end{align}
    where this step follows from Definition~\ref{def:Phi} and Definition~\ref{def:visual_decoder}.

    Besides, we also have:
    \begin{align}\label{eq:mu_func}
        \mu_t(\wt{u}) = & ~ H_N g(t) \notag \\
        = & ~ \Big( H_{N-1} ( I_s - \frac{1}{N} A )^\top + \frac{1}{N} \wt{u}_{N} B^\top \Big) g(t) \notag\\
        = & ~ \Bigg( H_{N-2} \Big ( (I_s - \frac{1}{N-1} A )^\top + \frac{1}{N-1} \wt{u}_{N} B^\top \Big) ( I_s - \frac{1}{N} A )^\top + \frac{1}{N} \wt{u}_{N} B^\top \Bigg) g(t) \notag\\
        = & ~ \Bigg( H_0 \prod_{\tau=1}^N (I_s - \frac{1}{\tau}A)^\top + \sum_{\tau=1}^N \Big( \prod_{\tau'=1}^{\tau - 1} (I_s - \frac{1}{\tau'} A )^\top\Big) \cdot \frac{1}{N+1-\tau} \wt{u}_{N+1-\tau} B^\top \Bigg) g(t)
    \end{align}
    where these steps follow from Definition~\ref{def:mu} and simple algebras.

    Recall $F_\theta(z, c, t) := f_{\cal T}( R([z^\top, c^\top, t]^\top) )$, we choose $n=1$, then there is a target function given by:
    \begin{align*}
        &  ~ f_{\cal T}([z^\top, c^\top, t]) \\
        = & ~ \frac{\sigma_t'(\wt{u})}{\sigma_t(\wt{u})} ( z - \Bigg( H_0 \prod_{\tau=1}^N (I_s - \frac{1}{\tau}A)^\top + \sum_{\tau=1}^N \Big( \prod_{\tau'=1}^{\tau - 1} (I_s - \frac{1}{\tau'} A )^\top\Big) \cdot \frac{1}{N+1-\tau} \wt{u}_{N+1-\tau} B^\top \Bigg) g(t)  ) \\
        & ~ + \Bigg( H_0 \prod_{\tau=1}^N (I_s - \frac{1}{\tau}A)^\top + \sum_{\tau=1}^N \Big( \prod_{\tau'=1}^{\tau - 1} (I_s - \frac{1}{\tau'} A )^\top\Big) \cdot \frac{1}{N+1-\tau} \wt{u}_{N+1-\tau}' B^\top \Bigg) g(t) \\
        & ~ + \Bigg( H_0 \prod_{\tau=1}^N (I_s - \frac{1}{\tau}A)^\top + \sum_{\tau=1}^N \Big( \prod_{\tau'=1}^{\tau - 1} (I_s - \frac{1}{\tau'} A )^\top\Big) \cdot \frac{1}{N+1-\tau} \wt{u}_{N+1-\tau} B^\top \Bigg) g'(t) \\
        = & ~ \frac{\sigma_t'(\wt{u})}{\sigma_t(\wt{u})} ( z \\ 
        & ~ - \Bigg( H_0 \prod_{\tau=1}^N (I_s - \frac{1}{\tau}A)^\top + \sum_{\tau=1}^N \Big( \prod_{\tau'=1}^{\tau - 1} (I_s - \frac{1}{\tau'} A )^\top\Big) \cdot \frac{1}{N+1-\tau} {\cal D}^{-1}\Big( (\Phi \wt{{\cal M}}(c))_{N+1-\tau} \Big) B^\top \Bigg) g(t)  ) \\
        & ~ + \Bigg( H_0 \prod_{\tau=1}^N (I_s - \frac{1}{\tau}A)^\top + \sum_{\tau=1}^N \Big( \prod_{\tau'=1}^{\tau - 1} (I_s - \frac{1}{\tau'} A )^\top\Big) \cdot \frac{1}{N+1-\tau} \Big({\cal D}^{-1}\Big( (\Phi \wt{{\cal M}}(c))_{N+1-\tau} \Big) \Big)' B^\top \Bigg) g(t) \\
        & ~ + \Bigg( H_0 \prod_{\tau=1}^N (I_s - \frac{1}{\tau}A)^\top + \sum_{\tau=1}^N \Big( \prod_{\tau'=1}^{\tau - 1} (I_s - \frac{1}{\tau'} A )^\top\Big) \cdot \frac{1}{N+1-\tau} {\cal D}^{-1}\Big( (\Phi \wt{{\cal M}}(c))_{N + 1 - \tau} \Big) B^\top \Bigg) g'(t)
    \end{align*}
    where the first step follows the combination of Theorem~\ref{thm:close_form} and Eq.~\eqref{eq:mu_func}, and the differentiablity of $\wt{u}_\tau$ is ensure by Assumption~\ref{ass:k}, the second step follows from Eq.~\eqref{eq:wt_u_func}.

    Following Theorem 2 and Theorem 3 in \cite{ybr+19}, we thus complete the proof by obtaining the theorem result.
\end{proof}

%% file: 12_app_interpolation.tex
\section{Interpolation and Extrapolation}
\label{sec:app:inter_extra}

This section first introduce properties of HiPPO-LegS in Appendix~\ref{sub:app:hippo_property}. Also, we bound the error of VLFM in Appendix~\ref{sub:app:error}.

\subsection{HiPPO-LegS Properties} \label{sub:app:hippo_property}

\begin{lemma}[Proposition 6 in \cite{gde+20}]\label{lem:optimal_projs}
    If the following conditions hold:
    \begin{itemize}
        \item Given a video caption distribution ${\cal V}_c$ as Definition~\ref{def:V_c}.
        \item For any $(V, c) \sim {\cal V}_c$, we define the discretized form of video as Definition~\ref{def:wt_V}.
        \item Let the observation matrix $\Phi: \{0, 1\}^{N \times \frac{T}{\Delta t}}$ be defined as Definition~\ref{def:Phi}.
        \item Let the visual decoder function $D: \R^d \rightarrow \R^D$ be defined as Definition~\ref{def:visual_decoder}.
        \item Let the ideal version of the sequence of latent patches $u \in \R^{\frac{T}{\Delta t} \times d}$ be defined as Definition~\ref{def:u}.
        \item Let the real-world version of the sequence of latent patches $\wt{u} \in \R^{N \times d}$ be defined as Definition~\ref{def:wt_u}.
        \item Let $H_N \in \R^{d \times s}$ be defined as Definition~\ref{def:H}.
        \item Let the function of polynomials $g(t)$ be defined as Definition~\ref{def:g_t}.
        \item Let the time-dependent mean of Gaussian distribution $\mu_t(\wt{u})$ be defined as Definition~\ref{def:mu}.
        \item Let the time-dependent standard deviation $\sigma_t(\wt{u})$ be defined as Definition~\ref{def:sigma}.
        \item Denote $\sigma_{\min} > 0$.
        \item Sample $z \sim \mathcal{N}(0, I_d)$.
        \item Define a model function $F_\theta: \R^d \times \R^\ell \times [0, T] \rightarrow \R^d$ with parameters $\theta$.
        \item Let the training objective ${\cal L}(\theta)$ be defined as Definition~\ref{def:L}.
        \item Let Assumptions~\ref{ass:k}, Assumption~\ref{ass:L_0}, Assumption~\ref{ass:M} and Assumption~\ref{ass:U} hold.
    \end{itemize}
    Then we have:
    \begin{align*}
        \| \mu_{\tau \cdot \Delta t}(\wt{u}) - \wt{u}_\tau \|_2 = O(t^{k}s^{-k+1/2})
    \end{align*}
\end{lemma}

\begin{proof}
    This lemma is a re-statement of Proposition 6 in \cite{gde+20}.
\end{proof}

\begin{lemma}[Proposition 3 in \cite{gde+20}]\label{lem:timescale_robustness}
    If the following conditions hold:
    \begin{itemize}
        \item Given a video caption distribution ${\cal V}_c$ as Definition~\ref{def:V_c}.
        \item For any $(V, c) \sim {\cal V}_c$, we define the discretized form of video as Definition~\ref{def:wt_V}.
        \item Let the observation matrix $\Phi: \{0, 1\}^{N \times \frac{T}{\Delta t}}$ be defined as Definition~\ref{def:Phi}.
        \item Let the visual decoder function $D: \R^d \rightarrow \R^D$ be defined as Definition~\ref{def:visual_decoder}.
        \item Let the ideal version of the sequence of latent patches $u \in \R^{\frac{T}{\Delta t} \times d}$ be defined as Definition~\ref{def:u}.
        \item Let the real-world version of the sequence of latent patches $\wt{u} \in \R^{N \times d}$ be defined as Definition~\ref{def:wt_u}.
        \item Let $H_N \in \R^{d \times s}$ be defined as Definition~\ref{def:H}.
        \item Let the function of polynomials $g(t)$ be defined as Definition~\ref{def:g_t}.
        \item Let the time-dependent mean of Gaussian distribution $\mu_t(\wt{u})$ be defined as Definition~\ref{def:mu}.
        \item Let the time-dependent standard deviation $\sigma_t(\wt{u})$ be defined as Definition~\ref{def:sigma}.
        \item Denote $\sigma_{\min} > 0$.
        \item Sample $z \sim \mathcal{N}(0, I_d)$.
        \item Define a model function $F_\theta: \R^d \times \R^\ell \times [0, T] \rightarrow \R^d$ with parameters $\theta$.
        \item Let the training objective ${\cal L}(\theta)$ be defined as Definition~\ref{def:L}.
        \item Let Assumptions~\ref{ass:k}, Assumption~\ref{ass:L_0}, Assumption~\ref{ass:M} and Assumption~\ref{ass:U} hold.
    \end{itemize}
    For any integer scale factor $\beta > 0$, the frames of video $\wt{V}_{\tau}$ is scaled to $\wt{V}_{\beta \tau}$, it doesn't affect the result of $H_N$ (Definition~\ref{def:H}).
\end{lemma}

\begin{proof}
    This lemma is a re-statement of Proposition 3 in \cite{gde+20}.
\end{proof}

\subsection{Error Bounds} \label{sub:app:error}

\begin{lemma}\label{lem:hippo_error}
    If the following conditions hold:
    \begin{itemize}
        \item Given a video caption distribution ${\cal V}_c$ as Definition~\ref{def:V_c}.
        \item For any $(V, c) \sim {\cal V}_c$, we define the discretized form of video as Definition~\ref{def:wt_V}.
        \item Let the observation matrix $\Phi: \{0, 1\}^{N \times \frac{T}{\Delta t}}$ be defined as Definition~\ref{def:Phi}.
        \item Let the visual decoder function $D: \R^d \rightarrow \R^D$ be defined as Definition~\ref{def:visual_decoder}.
        \item Let the ideal version of the sequence of latent patches $u \in \R^{\frac{T}{\Delta t} \times d}$ be defined as Definition~\ref{def:u}.
        \item Let the real-world version of the sequence of latent patches $\wt{u} \in \R^{N \times d}$ be defined as Definition~\ref{def:wt_u}.
        \item Let $H_N \in \R^{d \times s}$ be defined as Definition~\ref{def:H}.
        \item Let the function of polynomials $g(t)$ and matrix $G$ be defined as Definition~\ref{def:g_t}.
        \item Denote $1/\lambda^* := \lambda_{\min}(G) > 0$.
        \item Let the time-dependent mean of Gaussian distribution $\mu_t(\wt{u})$ be defined as Definition~\ref{def:mu}.
        \item Let the time-dependent standard deviation $\sigma_t(\wt{u})$ be defined as Definition~\ref{def:sigma}.
        \item Denote $\sigma_{\min} > 0$.
        \item Sample $z \sim \mathcal{N}(0, I_d)$.
        \item Define a model function $F_\theta: \R^d \times \R^\ell \times [0, T] \rightarrow \R^d$ with parameters $\theta$.
        \item Let the training objective ${\cal L}(\theta)$ be defined as Definition~\ref{def:L}.
        \item Let Assumptions~\ref{ass:k}, Assumption~\ref{ass:L_0}, Assumption~\ref{ass:M} and Assumption~\ref{ass:U} hold.
        \item $\delta \in (0, 1)$.
        \item Choosing $s = O(\frac{\Delta t}{T}\log((\frac{\Delta t}{T})^{1.5}/1/\lambda^*))$.
    \end{itemize}
    Particularly, we define:
    \begin{itemize}
        \item $\epsilon_1 := O(T^k s^{-k+1/2})$.
        \item $\epsilon_2 := O(\sqrt{d\log(d/\delta)})$.
        \item $\epsilon_3 := 1/\lambda^* U d^{0.5} \sqrt{\frac{T}{\Delta t} - N} \cdot \exp(O(\frac{T}{\Delta t}s))$.
    \end{itemize}
    Then with a probability at least $1 - \delta$, we have:
    \begin{align*}
        \| \psi_t(\wt{u}) - u_t \|_2 \leq \epsilon_1 + \epsilon_2 + \epsilon_3.
    \end{align*}
\end{lemma}

\begin{proof}
    We have:
    \begin{align*}
        \| \psi_t(\wt{u}) - u_t \|_2
        = & ~ \| \sigma_t(\wt{u}) \cdot z + \mu_t(\wt{u}) - u_t \|_2 \\
        \leq & ~ \| \sigma_t(\wt{u}) \cdot z \|_2 + \|  \mu_t(\wt{u}) - u_t \|_2 \\
        \leq & ~ \| z \|_2 + \|  \mu_t(\wt{u}) - u_t \|_2 \\
        \leq & ~ O( \sqrt{d \log(d/\delta)} ) +  \|  \mu_t(\wt{u}) - u_t \|_2 \\
        = & ~ \epsilon_2 + \|  \mu_t(\wt{u}) - u_t \|_2
    \end{align*}
    where the first step follows from Definition~\ref{def:psi}, the second step follows from triangle inequality, the third step follows from $\sigma_t(\wt{u}) \leq 1, \forall t \in [0, T]$ by some simple algebras and Definition~\ref{def:sigma}, the fourth step follows from the union bound of Gaussian tail bound (Fact~\ref{fac:gaussian_tail}), the last step follows from the definition of $\epsilon_2$.

    Then we get:
    \begin{align*}
        \|  \mu_t(\wt{u}) - u_t \|_2
        = & ~ \| H_N g(t) - u_t \|_2 \\
        = & ~ \| (M \cdot G)^\dagger (M \cdot u) \cdot g(t) - u_t \|_2 \\
        \leq & ~ \| (M \cdot G)^\dagger (M \cdot u) \cdot g(t) - G^\dagger u \cdot g(t) \|_2 + O((\frac{T}{\Delta t})^k s^{-k+1/2}) \\
        \leq & ~ \| ((M \cdot G)^\dagger (M \cdot u) - G^\dagger u \|_2 \cdot \| g(t) \|_2 + O((\frac{T}{\Delta t})^k s^{-k+1/2}) \\
        = & ~ \| ((M \cdot G)^\dagger (M \cdot u) - G^\dagger u \|_2 \cdot \| g(t) \|_2 + \epsilon_1
    \end{align*}
    where the first step follows from Definition~\ref{def:mu}, the second step follows from optimal error of solving $\| M G H - M u \|_2^2$, pesdueo-inverse matrix $(M \cdot G)^\dag \in \R^{d \times \frac{T}{\Delta t}}$ and defining a mask $M = \diag(m)$ where $m := \{0, 1\}^{\frac{T}{\Delta t}}$ and $\langle m, {\bf 1}_{\frac{T}{\Delta t}} \rangle = N$, the third step follows from the optimal error of solving $\| G H - u \|_2^2$, pesdueo-inverse matrix $G^\dag \in \R^{d \times \frac{T}{\Delta}}$ and Lemma~\ref{lem:optimal_projs}, the fourth step follows from Cauchy–Schwarz inequality and the last step follows from the definition of $\epsilon_2$.

    Next, we can show that:
    \begin{align*}
        \| (M \cdot G)^\dagger (M \cdot u) - G^\dagger u \|_2
        = & ~ \| (M \cdot G)^\dagger (M \cdot u) - G^\dagger (M \cdot u) + G^\dagger (M \cdot u) - G^\dagger u \|_2 \\
        \leq &~ \| (M \cdot G)^\dagger (M \cdot u)  - G^\dagger (M \cdot u) \|_2 + \| G^\dagger (M \cdot u) - G^\dagger u \|_2 \\
        \leq & ~ \| (M \cdot G)^\dagger - G^\dagger \|_2 \| (M \cdot u) \|_2 + \| G^\dagger \|_2 \| (M \cdot u) -  u \|_2
    \end{align*}
    where the first step follows from simple algebras, the second step follows from triangle inequality, the last step follows from Cauchy–Schwarz inequality.

    We first give:
    \begin{align}\label{eq:bound_G_dag}
        \| G^\dagger\|_2 \leq &~ 1/\lambda^* \sqrt{\frac{T}{\Delta t} \cdot s}
    \end{align}
    where this step follows from Definition~\ref{def:g_t}, Fact~\ref{fac:infity_norm_pesdueo_inverse} and the definition of $\ell_2$ norm.

    And:
    \begin{align*}
        \| u\|_2 \leq & ~ U \sqrt{\frac{T}{\Delta t} \cdot d}
    \end{align*}
    where this step follows from Assumption~\ref{ass:U} and the definition of $\ell_2$ norm.

    Also:
    \begin{align}\label{eq:bound_G}
        \| G\|_2 \leq & ~ \sqrt{\frac{T}{\Delta t} \cdot s} \exp( O(\frac{T}{\Delta t} \cdot s) ) 
    \end{align}
   where this step follows from Definition~\ref{def:g_t} and the definition of $\ell_2$ norm.

    Besides, we have:
    \begin{align*}
        \| (M \cdot G)^\dagger - G^\dagger \|_2
        \leq & ~ \frac{\| G^\dagger\|_2^2 \| I_{\frac{T}{\Delta t}} - M\|_2 \cdot \| G\|_2}{1 - \| G^\dagger\|_2 \cdot \| I_{\frac{T}{\Delta t}} - M\|_2 \cdot \| G\|_2} \\
        \leq & ~ \frac{{1/\lambda^*}^2 (\frac{T}{\Delta t} s)^{1.5} \sqrt{\frac{T}{\Delta t} - N} \cdot \exp(O(\frac{T}{\Delta t}s))}{1 - {1/\lambda^*} \frac{T}{\Delta t} s \sqrt{\frac{T}{\Delta t} - N}\cdot \exp(O(\frac{T}{\Delta t}s)) }
    \end{align*}
    where the first step follows from Fact~\ref{fac:pesdueo_inverse_diff}, simple algebras, and Cauchy–Schwarz inequality, the second step follows from Eq.~\eqref{eq:bound_G_dag}, Eq.~\eqref{eq:bound_G}, Definition~\ref{def:g_t} and simeple algebras.

    Combining all results, we get:
    \begin{align*}
        & ~ \| ((M \cdot G)^\dagger (M \cdot u) - G^\dagger u \|_2 \\
        \leq & ~ \frac{{1/\lambda^*}^2 (\frac{T}{\Delta t} s)^{1.5} \sqrt{\frac{T}{\Delta t} - N} \cdot \exp(O(\frac{T}{\Delta t}s))}{1 - {1/\lambda^*} \frac{T}{\Delta t} s \sqrt{\frac{T}{\Delta t} - N}\cdot \exp(O(\frac{T}{\Delta t}s)) } \cdot U \sqrt{\frac{T}{\Delta t} N d} + 1/\lambda^* \sqrt{\frac{T}{\Delta t} - N} \cdot U \sqrt{\frac{T}{\Delta t} \cdot d} \\
        \leq & ~ 1/\lambda^* U d^{0.5}  \sqrt{\frac{T}{\Delta t} (\frac{T}{\Delta t} - N)}  \cdot \Big( \frac{ {1/\lambda^*} (\frac{T}{\Delta t})^{1.5} N^{0.5} s^{1.5}  \cdot \exp(O(\frac{T}{\Delta t}s))}{1 - {1/\lambda^*} (\frac{T}{\Delta t})^{1.5} s \cdot \exp(O(\frac{T}{\Delta t}s))} + 1 \Big) \\
        \leq & ~ 1/\lambda^* U d^{0.5}  \sqrt{\frac{T}{\Delta t} (\frac{T}{\Delta t} - N)}  \cdot  \frac{ 1}{1 - {1/\lambda^*} (\frac{T}{\Delta t})^{1.5} s \cdot \exp(O(\frac{T}{\Delta t}s))}  \\
        \leq & ~ O\Big( 1/\lambda^* U d^{0.5}  \sqrt{\frac{T}{\Delta t} (\frac{T}{\Delta t} - N)}\Big)
    \end{align*}
    where the second and third steps follow from simple algebras, the last step follows from plugging the choice of $s$.

    Finally, we have:
    \begin{align*}
        \| ((M \cdot G)^\dagger (M \cdot u) - G^\dagger u \|_2 \cdot \| g(t) \|_2 
        \leq & ~ O\Big( 1/\lambda^* U d^{0.5}  \sqrt{\frac{T}{\Delta t} (\frac{T}{\Delta t} - N)}\Big) \cdot \sqrt{s} \exp(O(\frac{T}{\Delta t}s)) \\
        \leq & ~ 1/\lambda^* U d^{0.5} \sqrt{\frac{T}{\Delta t} - N} \cdot \exp(O(\frac{T}{\Delta t}s))  \\
        = & ~ \epsilon_3
    \end{align*}
    these steps follow from simple algebras, Definition~\ref{def:g_t} and the definition of $\epsilon_3$.
\end{proof}

\begin{theorem}\label{thm:inter_extra_polation}
    If the following conditions hold:
    \begin{itemize}
        \item Given a video caption distribution ${\cal V}_c$ as Definition~\ref{def:V_c}.
        \item For any $(V, c) \sim {\cal V}_c$, we define the discretized form of video as Definition~\ref{def:wt_V}.
        \item Let the observation matrix $\Phi: \{0, 1\}^{N \times \frac{T}{\Delta t}}$ be defined as Definition~\ref{def:Phi}.
        \item Let the visual decoder function $D: \R^d \rightarrow \R^D$ be defined as Definition~\ref{def:visual_decoder}.
        \item Let the ideal version of the sequence of latent patches $u \in \R^{\frac{T}{\Delta t} \times d}$ be defined as Definition~\ref{def:u}.
        \item Let the real-world version of the sequence of latent patches $\wt{u} \in \R^{N \times d}$ be defined as Definition~\ref{def:wt_u}.
        \item Let $H_N \in \R^{d \times s}$ be defined as Definition~\ref{def:H}.
        \item Let the function of polynomials $g(t)$ and matrix $G$ be defined as Definition~\ref{def:g_t}.
        \item Denote $1/\lambda^* := \lambda_{\min}(G) > 0$.
        \item Let the time-dependent mean of Gaussian distribution $\mu_t(\wt{u})$ be defined as Definition~\ref{def:mu}.
        \item Let the time-dependent standard deviation $\sigma_t(\wt{u})$ be defined as Definition~\ref{def:sigma}.
        \item Denote $\sigma_{\min} > 0$.
        \item Sample $z \sim \mathcal{N}(0, I_d)$.
        \item Define a model function $F_\theta: \R^d \times \R^\ell \times [0, T] \rightarrow \R^d$ with parameters $\theta$.
        \item Let the training objective ${\cal L}(\theta)$ be defined as Definition~\ref{def:L}.
        \item Let Assumptions~\ref{ass:k}, Assumption~\ref{ass:L_0}, Assumption~\ref{ass:M} and Assumption~\ref{ass:U} hold.
        \item $\delta \in (0, 1)$.
    \end{itemize}
    Particularly, we define:
    \begin{itemize}
        \item $\epsilon_1 := O(T^k s^{-k+1/2})$.
        \item $\epsilon_2 := O(\sqrt{d\log(d/\delta)})$.
        \item $\epsilon_3 := 1/\lambda^* U d^{0.5} \sqrt{\frac{T}{\Delta t} - N} \cdot \exp(O(\frac{T}{\Delta t}s))$.
    \end{itemize}
    Then with a probability at least $1 - \delta$, we have:
    \begin{align*}
        \| {\cal D}(z + \int_0^{t} F_\theta(z, c, t') \d t') - u_t\|_2 \leq \epsilon_0 + L_0 (\epsilon_1 + \epsilon_2 + \epsilon_3).
    \end{align*}
\end{theorem}

\begin{proof}
    This proof follows from the combination of Assumption~\ref{ass:L_0}, Theorem~\ref{thm:uat} and Lemma~\ref{lem:hippo_error}.
\end{proof}